%% file: icml2022.tex
\documentclass[nohyperref]{article}

\usepackage{microtype}
\usepackage{graphicx}
\usepackage{subfigure}
\usepackage{booktabs} % for professional tables
\usepackage{hyperref}

\usepackage[linesnumbered,ruled,vlined,boxed]{algorithm2e}
\usepackage[accepted]{icml2022}

\usepackage{amsmath}
\usepackage{amssymb}
\usepackage{mathtools}
\usepackage{amsthm}

\input{preamble}
\usepackage{wrapfig, framed, caption}
\usepackage{multirow}
\hypersetup{colorlinks,linkcolor={blue},citecolor={blue},urlcolor={black}}

\def\Dista{A}
\def\Distb{B}
\def\facA{A_1}
\def\facAA{A_2}
\def\facB{B_1}
\def\facBB{B_2}

\usepackage[capitalize,noabbrev]{cleveref}

\usepackage[textsize=tiny]{todonotes}

\icmltitlerunning{Linear-Time Gromov Wasserstein Distances using Low Rank Couplings and Costs}

\begin{document}

\twocolumn[
\icmltitle{Linear-Time Gromov Wasserstein Distances\\ using Low Rank Couplings and Costs}




\begin{icmlauthorlist}
\icmlauthor{Meyer Scetbon}{crest}
\icmlauthor{Gabriel Peyré}{ens}
\icmlauthor{Marco Cuturi}{crest-2}
\end{icmlauthorlist}

\icmlaffiliation{crest}{CREST-ENSAE}
\icmlaffiliation{ens}{CNRS and ENS, PSL}
\icmlaffiliation{crest-2}{CREST-ENSAE, work partly done at Google, currently at Apple}

\icmlcorrespondingauthor{meyer scetbon}{meyer.scetbon@ensae.fr}
% \icmlcorrespondingauthor{marco cuturi}{cuturi@apple.com}

\icmlkeywords{Machine Learning, ICML}

\vskip 0.3in
]



\printAffiliationsAndNotice{}  % leave blank if no need to mention equal contribution

\begin{abstract}
The ability to align points across two related yet incomparable point clouds (e.g. living in different spaces) plays an important role in machine learning. 
The Gromov-Wasserstein (GW) framework provides an increasingly popular answer to such problems, by seeking a low-distortion, geometry-preserving assignment between these points.
As a non-convex, quadratic generalization of optimal transport (OT), GW is NP-hard. While practitioners often resort to solving GW approximately as a nested sequence of entropy-regularized OT problems, the cubic complexity (in the number $n$ of samples) of that approach is a roadblock.
We show in this work how a recent variant of the OT problem that restricts the set of admissible couplings to those having a low-rank factorization is remarkably well suited to the resolution of GW:
when applied to GW, we show that this approach is not only able to compute a stationary point of the GW problem in time $O(n^2)$, but also uniquely positioned to benefit from the knowledge that the initial cost matrices are low-rank, to yield a linear time $O(n)$ GW approximation. Our approach yields similar results, yet orders of magnitude faster computation than the SoTA entropic GW approaches, on both simulated and real data. 
\end{abstract}

\input{sections/intro}
\input{sections/background}

\input{sections/methods}
\input{sections/methods2}

\input{sections/methods3}
\input{sections/experiment}

\newpage
\clearpage
\bibliography{biblio}
\bibliographystyle{plainnat}

%%%%%%%%%%%%%%%%%%%%%%%%%%%%%%%%%%%%%%%%%%%%%%%%%%%%%%%%%%%%%%%%%%%%%%%%%%%%%%%
%%%%%%%%%%%%%%%%%%%%%%%%%%%%%%%%%%%%%%%%%%%%%%%%%%%%%%%%%%%%%%%%%%%%%%%%%%%%%%%
% APPENDIX
%%%%%%%%%%%%%%%%%%%%%%%%%%%%%%%%%%%%%%%%%%%%%%%%%%%%%%%%%%%%%%%%%%%%%%%%%%%%%%%
%%%%%%%%%%%%%%%%%%%%%%%%%%%%%%%%%%%%%%%%%%%%%%%%%%%%%%%%%%%%%%%%%%%%%%%%%%%%%%%
\clearpage
\appendix
\input{sections/suppmat}

\end{document}

%% file: preamble.tex
\pdfoutput=1

\newtheorem{defn}{Definition}
\newtheorem{prop}{Proposition}

\newtheorem{assumption}{Assumption}

\DeclareMathOperator{\Diag}{diag}

\DeclareMathOperator*{\argmin}{argmin}

\newcommand{\eqdef}{:=}
\definecolor{mygreen}{rgb}{0.0, 0.7, 0.0}
\newcommand{\R}{\mathbb{R}}

%% file: sections/intro.tex
\section{Introduction}
\vskip-.2cm
\begin{figure}
    \centering
    \includegraphics[width=0.45\textwidth]{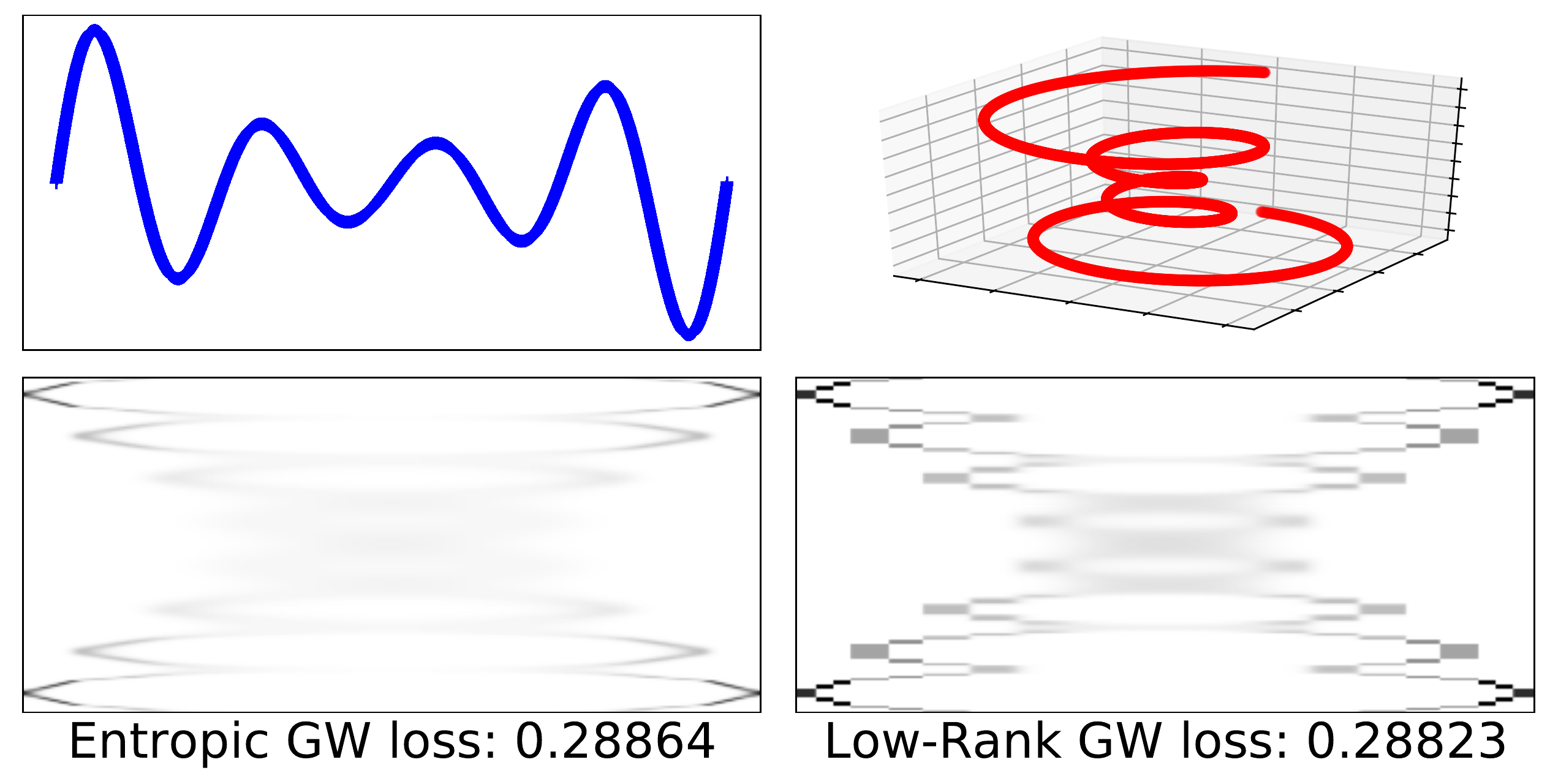}
    \caption{\emph{Top row:} Two curves in 2D and 3D, with $n=5000$ points. \emph{Bottom row:} coupling and GW loss obtained with the SoTA $O(n^3)$ entropic approach~\citep{peyre2016gromov} (left) and with our linear $O(n)$ method (right) when using the squared Euclidean distances as the ground costs. 
}\label{fig-GT-coupling}
\vskip-.6cm
\end{figure}
\textbf{Increasing interest for Gromov-Wasserstein... }
Several problems in machine learning require comparing datasets that live in heterogeneous spaces. This situation arises typically when realigning two distinct views (or features) from points sampled from similar sources. Recent applications to single-cell genomics~\citep{SCOT2020,blumberg2020mrec} %and NLP~\cite{conneau2017word,alaux2018unsupervised} 
provide a case in point: Thousands of cells taken from the same tissue are split in two groups, each processed with a different experimental protocol, resulting in two distinct sets of heterogeneous feature vectors; 
%Thousands of word embeddings for two languages are learned independently. In both cases, 
Despite this heterogeneity, one expects to find a mapping registering points from the first to the second set, since they contain similar overall information.
That realignment is usually carried out using the Gromov-Wasserstein (GW) machinery proposed by~\citet{memoli2011gromov} and ~\citet{sturm2012space}. GW seeks a relaxed assignment matrix that is as close to an isometry as possible, as quantified by a quadratic score.
GW has practical appeal: It has been used in supervised learning~\citep{xu2019gromov}, generative modeling~\citep{bunne2019learning}, domain adaptation \citep{chapel2020partial}, 
structured prediction \citep{vayer2018fused}, quantum chemistry \citep{peyre2016gromov} and 
alignment layers~\citep{ezuz2017gwcnn}. 

\textbf{...despite being hard to solve. }
Since GW is an NP-hard problem, all applications above rely on heuristics, the most popular being the sequential resolution of nested entropy-regularized OT problems. That approximation remains costly, requiring $\mathcal{O}(n^3)$ operations when dealing with two datasets of $n$ samples.
Our goal is to reduce that complexity, by exploiting and/or enforcing low-rank properties of matrices arising \textit{both} in data and variables of the GW problem.
%We show below that this low rank approach allows to get rid of several limiting factors when computing GW, going as far as potentially turning it from a cubic to a linear time complexity algorithm.

\begin{figure}
    \centering
    \includegraphics[width=0.45\textwidth]{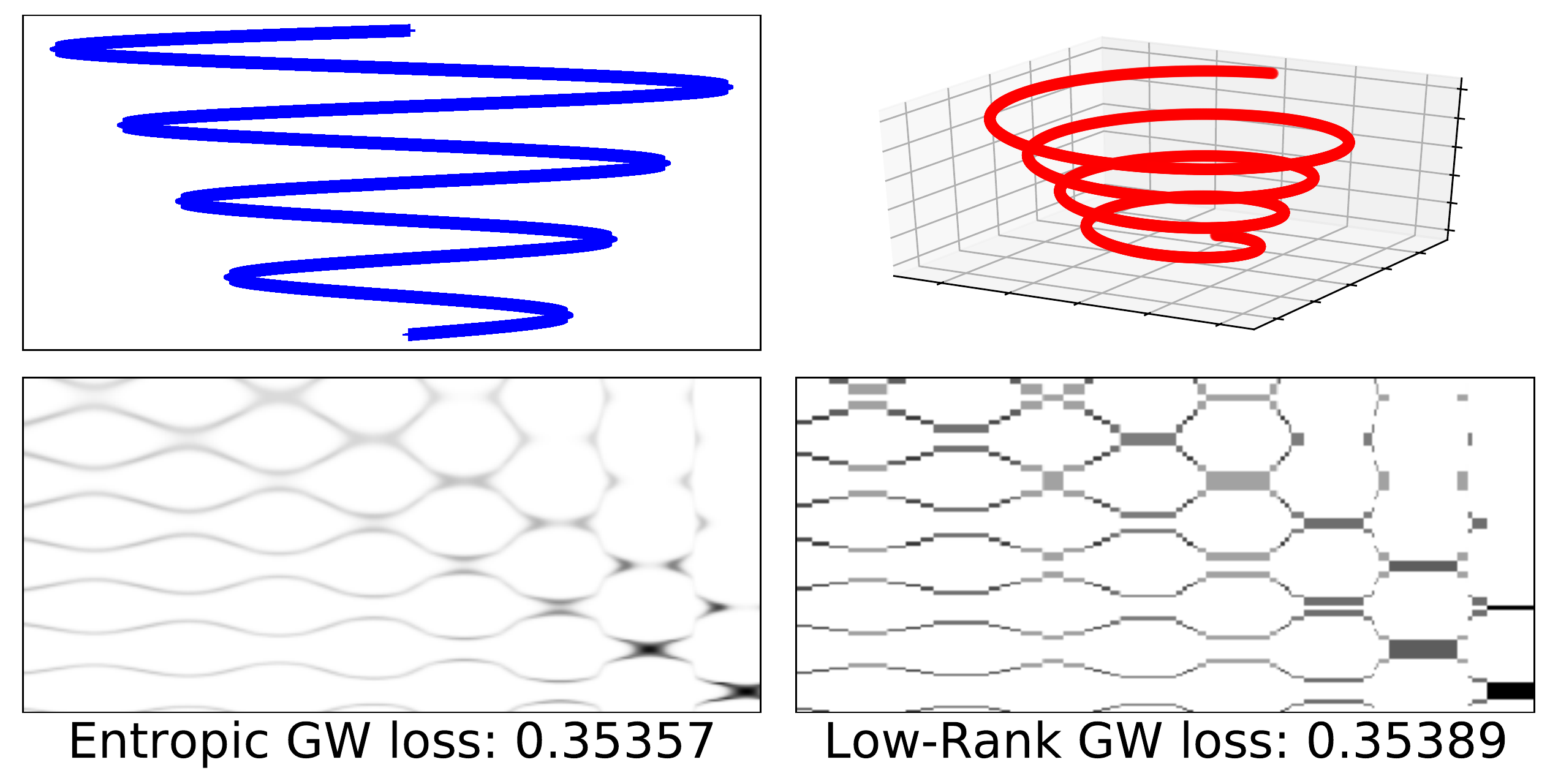}
    \caption{\emph{Top row:} Two curves in 2D and 3D, with $n=5000$ points. \emph{Bottom row:} coupling and GW loss obtained with the SoTA $O(n^3)$ entropic approach~\citep{peyre2016gromov} (left) and with our linear $O(n)$ method (right) when using the squared Euclidean distance as the ground cost for both point clouds. See Appendix~\ref{sec-illusation-fig1} for more details.
}\label{fig-GT-coupling-1}
\vskip-.5cm
\end{figure}

%%%%%%%%%%%%%%%%%%%%%%%%%%%%%%%%%%%%%%%%%%%%%%%%%%%%%%%%%%%%%%%%%%%%%%%%%%%
\textbf{OT: from cubic to linear complexity. }
Compared to GW, aligning two populations embedded in the \textit{same} space is far simpler, and corresponds to the usual optimal transport (OT) problem~\citep{Peyre2019computational}.%, which has found numerous applications, in NLP~\citep{kusner2015word}, single cell tracking~\citep{schiebinger2019optimal} or neuro-imaging~\citep{janati2020multi} for instance.
Given a $n\times m$ cost matrix $C$ and two marginals, the OT problem minimizes $\mathcal{L}_C(P):=\langle C, P\rangle$ w.r.t. a coupling matrix $P$ satisfying these marginal constraints. For computational and statistical reasons, most practitioners rely on regularized approaches $\mathcal{L}^\varepsilon_C(P):=\langle C, P\rangle + \varepsilon\text{reg}(P)$. When $\text{reg}$ is the neg-entropy, \citeauthor{Sinkhorn64}'s algorithm can be efficiently employed~\citep{cuturi2013sinkhorn,altschuler2017near,lin2019efficient}.
%, which is better suited for ML applications (parallelism, GPU execution, differentiability)~\citep{
% DO WE NEED WHAT'S BELOW IN THE INTRO?
% The Sinkhorn algorithm outputs an $\epsilon$-approximation in $O(n^2/\epsilon^3)$ operations~\citep{altschuler2017near} of the LP solution. 
%\mc{not sure we need that granularity?}
%
The Sinkhorn iteration has $O(nm)$ complexity, but this can be sped-up using either a low-rank factorizations (or approximations) of the \textit{kernel} matrix $K:=e^{-C/\varepsilon}$~\citep{2015-solomon-siggraph,altschuler2018massively,altschuler2018approximating,scetbon2020linear}, or, alternatively and as proposed by~\citet{scetbon2021lowrank,forrow2019statistical}, by imposing a low-rank \textit{constraint} on the coupling $P$. A goal in this paper is to show that this latter route is remarkably well suited to the GW problem.

%%%%%%%%%%%%%%%%%%%%%%%%%%%%%%%%%%%%%%%%%%%%%%%%%%%%%%%%%%%%%%%%%%%%%%%%%%%
\textbf{GW: from NP-hard to cubic approximations. }
The GW problem replaces the linear objective in OT by a \textit{non-convex}, \textit{quadratic}, objective $\mathcal{Q}_{A,B}(P):= \text{cst}-2\langle APB,P\rangle$ parameterized by \textit{two} square cost matrices $A$ and $B$. Much like OT is a relaxation of the optimal assignment problem, GW is a relaxation of the quadratic assignment problem (QAP). Both GW and QAP are NP-hard~\citep{burkard1998quadratic}. In practice, linearizing iteratively $\mathcal{Q}_{A,B}$ works well~\citep{gold1996softassign,solomon2016entropic}: recompute a synthetic cost $C_{t}:=AP_{t-1}B$, use Sinkhorn to get $P_{t}:= \argmin_P \langle C_t, P\rangle + \varepsilon\text{reg}(P)$, repeat.
%
% \begin{wrapfigure}{r}{0.5\textwidth}
% \vskip-.2cm
% \begin{minipage}{0.5\textwidth}
% \centering
% \includegraphics[width=1\textwidth]{figures/Ground_Truth/plot_coupling_results.pdf}
% \caption{\emph{Top row:} we compute the GW coupling between two curves in 2D and 3D, with $n=m=10000$ points. These points are endowed with the squared L2 distance. \emph{Bottom row:} coupling obtained with the SoTA entropic approach~\cite{gold1996softassign,peyre2016gromov}, compared with our linear method with rank $r=10$. See Appendix~\ref{sec-illusation-fig1} for more details.
% }\label{fig-GT-coupling}
% \end{minipage}
% \end{wrapfigure}
%
This is akin to a mirror-descent scheme~\citep{peyre2016gromov}, %and in the special case of Euclidean distance matrices, the loss is concave 
interpreted as a bi-linear relaxation in certain cases~\citep{konno1976maximization}.
% bilinear concave programming explains why it works \citep{konno1976maximization} alternative explanation is via mirror descent \citep{peyre2016gromov}
%

\textbf{Challenges to speed up GW. }
Several obstacles stand in the way of speeding up GW. The re-computation of $C_{t}=AP_{t-1}B$ at each outer iteration is an issue, since it requires $O(n^3)$ operations~\citep[Prop. 1]{peyre2016gromov}. We only know of two broad approaches that achieve tractable running times: \textit{(i)} Solve related, yet significantly different, proxies of the GW energy, either by embedding points \textit{as} univariate measures~\citep{memoli2011gromov,sato2020fast}, by using a sliced mechanism when restricted to Euclidean settings~\citep{vayer2019sliced} or by considering tree metrics for supports of each probability measure~\cite{le2021flow}, \textit{(ii)} Reduce the size of the GW problem through quantization of input measures~\citep{chowdhury2021quantized}.
%which results in an approximation that retains metric properties and outputs an upper-bound on the GW distance. 
%
or recursive clustering approaches~\citep{xu2019scalable,blumberg2020mrec}). Interestingly, no work has, to our knowledge, tried yet to accelerate Sinkhorn iterations withing GW.

\textbf{Our contributions: a quadratic to linear GW approximation.}
Our method addresses the problem by taking the GW as it is, overcoming limitations that may arise from a changing cost matrix $C_t$.
We show first that a low-rank factorization (or approximation) of the two input cost matrices that define GW, one for each measure, can be exploited to lower the complexity of recomputing $C_t$ from cubic to quadratic.
We show next, independently, that using the low-rank approach for \textit{couplings} advocated by~\citet{scetbon2021lowrank} to solve OT can be inserted in the GW pipeline and result in a $O(n^2)$ strategy for GW, with no prior assumption on input cost matrices. We also briefly explain why methods that exploit the geometrical properties of $C$ (or its kernel $K=e^{-C}$) to obtain faster iterations are of little use in a GW setup, because of the necessity to re-instantiate a new cost $C_t$ at each outer iteration.
% This method works hand-in-hand with entropic regularization and leads to a Sinkhorn-like algorithm.
%
Finally, we show that both low-rank assumptions (on costs and couplings) can be combined to shave yet another factor and reach GW approximation with linear complexity in time and memory.
We provide experiments, on simulated and real datasets, which show that our approach has comparable performance to entropic-regularized GW and its practical ability to reach ``good'' local minima to GW, for a considerably cheaper computational price, and with a conceptually different regularization path (see  Fig.~\ref{fig-GT-coupling},\ref{fig-GT-coupling-1}), yet can scale to millions of points.

% Low rank coupling makes it $O(n)$

% \textbf{Aligning Incomparable Spaces} When comparing two families of points of respective sizes $n$ and $m$, the GW approach shares several modelling assumptions with the ``simpler'' optimal transport (OT) problem~\cite{Peyre2019computational} but is considerably harder to solve, because its objective is quadratic and non-convex in the space of $n\times m$ couplings (therefore, only \textit{evaluating} that objective would incur a cost $\mathcal{O}(n^2m^2)$ in the most general setting. Unlike OT, the goal is therefore not to reach provably an optimal solution with a given computational budget, but rather to propose useful heuristics that strike a good balance between faithfulness to that original objective, stability and scalability. 

% \textbf{Reduction to linear OT} A popular heuristic to decrease the GW objective is to use linearization of that quadratic objective, and solve the resulting linear problem using a standard OT solver. Still, computing 

%% file: sections/background.tex
\section{Background on Gromov-Wasserstein}
\textbf{Comparing measured metric spaces.}~Let $(\mathcal{X},d_{\mathcal{X}})$ and  $(\mathcal{Y},d_{\mathcal{Y}})$ be two metric spaces, and $\mu:=\sum_{i=1}^n a_i\delta_{x_i}$ and $\nu:=\sum_{i=j}^m b_j\delta_{y_j}$ two discrete probability measures, where $n,m\geq 1$; $a$, $b$ are probability vectors in the simplicies $\Delta_n, \Delta_m$ of size $n$ and $m$; and $(x_1,\dots,x_n)$, $(y_1,\dots,y_m)$ are families in $\mathcal{X}$ and $\mathcal{Y}$. Given $q\geq 1$, the following square pairwise \textit{cost} matrices encode the geometry \textit{within} $\mu$ and $\nu$,
$$\Dista:=[d_{\mathcal{X}}^q(x_i,x_{i'})]_{1 \leq i,i'\leq n},\; \Distb:=[d_{\mathcal{Y}}^q(x_j,x_{j'})]_{1 \leq i,i'\leq m}\,$$ 
The GW discrepancy between these two discrete metric measure spaces $(\mu, d_\mathcal{X})$ and $(\nu, d_\mathcal{Y})$ is the solution of the following non-convex quadratic problem, written for simplicity as a function of $(a,\Dista)$ and $(b,\Distb)$:
\begin{align}
\label{eq-GW}
    &\text{GW}((a, \Dista),(b, \Distb))=\!\!\min_{P\in\Pi_{a,b}}\!\!\mathcal{Q}_{\Dista, \Distb}(P), \,\\ &\text{where } \Pi_{a,b} \eqdef \{P\in\mathbb{R}_{+}^{n\times m} | P\mathbf{1}_m=a, P^{T}\mathbf{1}_n=b\},\nonumber 
\end{align}
and the energy $\mathcal{Q}_{\Dista, \Distb}$ is a quadratic function of $P$ designed to measure the distortion of the assignment:
\begin{align}
\label{eq-obj-GW}
\mathcal{Q}_{\Dista, \Distb}(P)\eqdef \sum_{i,j,i',j'} (\Dista_{i,i'} - \Distb_{j,j'})^2 P_{i,j}P_{i',j'}\;.
\end{align}
\citet{memoli2011gromov} proves that $\text{GW}^{\tfrac{1}{2}}$ defines a distance on the space of metric measure spaces quotiented by measure-preserving isometries. ~\eqref{eq-obj-GW} can be evaluated in $\mathcal{O}(n^2m+nm^2)$ operations, rather than using $n^2m^2$ terms:
\begin{align}
\label{reformulation-GW-obejective}
  \mathcal{Q}_{\Dista, \Distb}(P)=\langle \Dista^{\odot2} a,a\rangle +  \langle \Distb^{\odot2}b,b\rangle -2\langle \Dista P\Distb,P\rangle\; ,
\end{align}
where $\odot$ is the Hadamard (elementwise) product or power.

% \begin{algorithm}[H]
% \SetAlgoLined
% \textbf{Inputs:} $\color{red}D\color{black},\color{blue}D',\color{red}a\color{black},\color{blue}d\color{black},\color{brown}r\color{black}$\\
% \For{$t=1,\dots$}{
%     $\color{purple}C_1\color{black}\gets -\color{red}D\color{black} Q\Diag(1/g)$\\
%     $\color{purple}C_2\color{black}\gets R^T\color{blue}D'\color{black}$\\
%     $\color{green}K^{(1)}\color{black}\gets Q\odot\exp(4\gamma \color{purple}C_1C_2\color{black} R \Diag(1/g))$\\ 
%     $\color{green}K^{(2)}\color{black}\gets R\odot\exp(4\gamma (\color{purple}C_1C_2\color{black})^TQ \Diag(1/g))$\\
%     $\omega\gets \mathcal{D}(Q^T\color{purple}C_1C_2\color{black}R)$\\
%     $\color{green}K^{(3)}\color{black}\gets g\odot \exp(-4\gamma\omega/g^2)$\\
%     $Q,R,g\gets \argmin\limits_{{\mathbf{\zeta}\in\mathcal{C}_1(\color{red}a\color{black},\color{blue}b\color{black},\color{brown}r\color{black})\cap \mathcal{C}_2(\color{brown}r\color{black})}}\text{KL}(\mathbf{\zeta},\color{green}\textbf{K}_{k}\color{black})$
%   }
% \textbf{Return:} $\mathcal{Q}_{\color{red}D\color{black},\color{blue}D'\color{black}}(Q\Diag(1/g)R^T)$
% \caption{Low-Rank GW \label{alg-MDGW-LR}
% % , $\text{GW-LR}_{\varepsilon,\alpha}^{(r)}((a,\Dista), (b,\Distb))$ \label{alg-MDGW-LR}
% }
% \end{algorithm}

\textbf{Entropic Gromov-Wasserstein.} The original GW problem~\eqref{eq-GW} can be regularized using entropy ~\citep{gold1996softassign,solomon2016entropic}, leading to problem:
\begin{align}
\label{eq-GW-ent}
    \text{GW}_{\varepsilon}((a, \Dista),(b, \Distb))=\min_{P\in\Pi_{a,b}}\mathcal{Q}_{\Dista, \Distb}(P) -\varepsilon H(P)\,,
\end{align}
where $H(P):=-\sum_{i,j}P_{i,j}(\log(P_{i,j})-1)$ is $P$'s entropy. \citet{peyre2016gromov} propose to solve that problem using mirror descent (MD), w.r.t. the $\text{KL}$ divergence. Their algorithm boils down to solving a sequence of regularized OT problems, as in Algo.~\ref{alg-mirror-descent}: Each KL projection in Line~\ref{line:KL} is solved efficiently with the Sinkhorn algorithm~\citep{cuturi2013sinkhorn}.
% \begin{wrapfigure}{r}{0.4\textwidth}
%\vskip-.4cm
% \begin{minipage}{0.4\textwidth}
% \centering

\begin{algorithm}
\SetKwInOut{Input}{Input}
\Input{$a\in\Delta_n, \Dista \in\R^{n\times n}\!\!, b\in\Delta_m, \Distb \in\R^{m\times m}\!\!, \varepsilon>0$}
$P = ab^T$\quad{\color{red}{\texttt{nm}}}\\
\For{$t=0,\dots$}{
    $C \gets - 4 \Dista P \Distb$ \quad{\color{violet}{\texttt{nm(n+m)}}} \label{line:updC}\\
    $K_\varepsilon \gets \exp(-C/\varepsilon)$\quad{\color{red}{\texttt{nm}}}\\
    $P\gets \argmin\limits_{P\in\Pi(a,b)} \text{KL}(P,K_\varepsilon)$\quad{\color{red}{$\mathcal{O}$(\texttt{nm})}}\label{line:KL}}
$\text{GW} =\mathcal{Q}_{\Dista, \Distb}(P)$  \,{\color{violet}{\texttt{nm(n+m)}}}\label{line:eval}\\
\KwResult{\text{GW}}
\caption{Entropic-GW \label{alg-mirror-descent}}
\end{algorithm} 
% % \end{minipage}
% % \vskip-.2cm
% % \end{wrapfigure}

\textbf{Computational complexity.} Given a cost matrix $C$, the KL projection of $K_{\varepsilon}$ onto the polytope $\Pi(a,b)$, where $\text{KL}(P,Q)=\langle P, \log(P/Q)-1\rangle$, is carried out in Line \ref{line:KL} of the inner loop of Algo.~\ref{alg-mirror-descent} using the Sinkhorn algorithm, through matrix-vector products. 
This quadratic complexity (in {\color{red}{\textbf{red}}}) is dominated by the cost of updating matrix $C$ at each iteration in Line~\ref{line:updC}, which requires $\mathcal{O}(n^2m+nm^2)$ algebraic operations (cubic, in {\color{violet}{\textbf{violet}}}).
As noted above, evaluating the objective $\mathcal{Q}_{\Dista, \Distb}(P)$ in Line~\ref{line:eval} is also cubic. 

\textbf{Step-by-step guide to reaching linearity.} We show next in \S\ref{sec-LR-cost} that these iterations can be sped up when the distance matrices are low-rank (or have low-rank approximations), in which case the cubic updates in $C$ and evaluation of $\mathcal{Q}_{\Dista, \Distb}$ in Lines~\ref{line:updC},~\ref{line:eval} become quadratic. Independently, we show in \S\ref{sec:imposing} that, with \emph{no assumption} on these cost matrices, replacing the Sinkhorn call in Line~\ref{line:KL} with a low-rank approach~\citep{scetbon2021lowrank} can lower the cost of Lines~\ref{line:updC},~\ref{line:eval} to quadratic (while also making Line~\ref{line:KL} linear). Remarkably, we show in \S\ref{sec-lin-GW} that these two approaches can be combined in Lines~\ref{line:updC},~\ref{line:eval}, to yield, to the best of our knowledge, the first linear time/memory algorithm able to match the performance of the Entropic-GW approach.

%% file: sections/methods.tex
\section{Low-rank (Approximated) Costs}
\label{sec-LR-cost}
\textbf{Exact factorization for distance matrices.} consider
\begin{assumption}
\label{assump-low-rank}
$\Dista$ and $\Distb$ admit a low-rank factorization: there exists $\facA,\facAA\in\mathbb{R}^{n\times d}$ and $\facB, \facBB\in\mathbb{R}^{m\times d'}$ s.t.
$\Dista = \facA\facAA^{T}$ and $\Distb = \facB\facBB^{T}$, where $d\ll n, d'\ll m$.
\end{assumption}

A case in point is when both $\Dista$ and $\Distb$ are \textit{squared} Euclidean distance matrices, with a sample size that is much larger than ambient dimension. This case is highly relevant in practice, since it covers most applications of OT to ML. Indeed, the $d\ll n$ assumption usually holds, since cases where $d\gg n$ fall in the ``curse of dimensionality'' regime where OT is less useful~\citep{dudley1966weak,weed2017sharp}. Writing $X= [x_1,\dots,x_n]\in\mathbb{R}^{d\times n}$, if $\Dista = \left[\Vert x_i-x_j\Vert_2^2\right]_{i,j}$, then one has, writing $z=(X^{\odot2})^T\mathbf{1}_d \in\mathbb{R}^n$ that $\Dista=z\mathbf{1}_n^T + \mathbf{1}_n z^T - 2 X^T X.$ Therefore by denoting $\facA=[z,\mathbf{1}_n,-\sqrt{2}X^T]\in\mathbb{R}^{n\times (d+2)}$ and $\facAA=[\mathbf{1}_n,z,\sqrt{2}X^T]\in\mathbb{R}^{n\times (d+2)}$ we obtain the factorization above.
% \begin{rmq}
% Note that in the GW distance $D$ and $\Distb$ are always distance matrices and therefore a low-rank approximation can be obtained in linear time as discussed in Section~\ref{sec-low-rank-distance}.
% % Indeed, by denoting $r_{\text{tot}}:=\sum_{k=1}^K r_k$, one can define $A_{\text{tot}}:=[A_1,...,A_K]\in\mathbb{R}^{n\times r_{\text{tot}}}$ and $B_{\text{tot}}:=[B_1,...,B_K]\in\mathbb{R}^{n\times r_{\text{tot}}}$ and obtain that $D=A_{\text{tot}}B_{\text{tot}}^T$. 
% \end{rmq}
Under Assumption~\ref{assump-low-rank}, the complexity of Algo.~\ref{alg-mirror-descent} is reduced to  $O(n^2)$: Line~\ref{line:updC} reduces to:
\begin{align*}
    C  = - 4 \facA \facAA^{T} P \facB \facBB^{T}\,,
\end{align*}
in $nm(d+d') + dd'(n+m)$ algebraic operations, while Line~\ref{line:eval}, using the reformulation of $\mathcal{Q}_{\Dista,\Distb}(P)$ in~\eqref{reformulation-GW-obejective}, becomes quadratic as well. Indeed, writing $G_1
:=\facA^T P\facBB$ and $G_2
:=\facAA^T P\facB$, both in $\mathbb{R}^{d\times d'}$, one has $\langle \Dista P\Distb ,P\rangle =\mathbf{1}_d^T(G_1 \odot G_2)\mathbf{1}_{d'}$.
Computing $G_1, G_2$ given $P$ requires only $2(nmd+mdd')$, and computing their dot product adds $dd'$ algebraic operations. The overall complexity to compute  $\mathcal{Q}_{\Dista,\Distb}(P)$ is $\mathcal{O}(nmd+mdd')$.

\begin{algorithm}
\SetAlgoLined
\textbf{Inputs:} $\facA,\facAA\in\mathbb{R}^{n\times d},\facB,\facBB,\in\mathbb{R}^{m\times d'} a,b,\varepsilon$\\
$P=ab^T$\,\,\,\,\,{\color{red}{\texttt{nm}}}\\
\For{$t=0,\dots$}{
$G_2 \gets \facAA^{T} P \facB$ \,\,\,\,\,{\color{red}{\texttt{nmd + mdd'}}}\\
$C \gets - 4 \facA G_2 \facBB^{T}$  \,\,\,\,\,{\color{red}{\texttt{nmd' + ndd'}}}\\
$K_\varepsilon \gets \exp(-C/\varepsilon)$\,\,\,\,\,{\color{red}{\texttt{nm}}}\\
$P\gets \argmin\limits_{P\in\Pi(a,b)} \text{KL}(P,K_\varepsilon)$\,\,\,\,\,{\color{red}{$\mathcal{O}$(\texttt{nm})
  }} }
 $c_1\gets a^T(\facA\facAA^T)^{\odot2}a + b^T(\facB\facBB^T)^{\odot2}b$\label{line-step-9-alg-2} \,\,\,{\color{red}{\texttt{n²d'+m²d'}}}\\
  $G_2 \gets \facAA^{T} P \facB$ \,\,\,\,\,{\color{red}{\texttt{nmd + mdd'}}}\\
  $G_1 \gets \facA^{T} P \facBB$\,\,\,\,\,{\color{red}{\texttt{nmd + mdd'}}} \\
  $c_2\gets - 2 \mathbf{1}_d^T(G_1 \odot G_2)\mathbf{1}_{d'}$\,\,\,\,\,{\color{mygreen}{\texttt{dd'}}}\\
 $\mathcal{Q}_{\Dista,\Distb}(P)\gets c_1 + c_2$\\
 \textbf{Return: } $\mathcal{Q}_{\Dista,\Distb}(P)$
\caption{Quadratic Entropic-GW \label{alg-quad-GW}}
\end{algorithm}
% \end{minipage}\vskip-.2cm
% \end{wrapfigure}

\textbf{General distance matrices.} When the original cost matrices $\Dista, \Distb$ are not low-rank but describe distances, we build upon recent works that output their low-rank approximation in linear time~\citep{bakshi2018sublinear,indyk2019sampleoptimal}. These algorithms produce, for any distance matrix $A\in\mathbb{R}^{n\times m}$ and $\tau>0$, matrices $A_1\in\mathbb{R}^{n\times d}$, $A_2\in\mathbb{R}^{m\times d}$ in $\mathcal{O}((m+n)\text{poly}(\frac{d}{\tau}))$ operations such that, with probability at least $0.99$,
\begin{align*}
    \Vert A - A_1A_2^T\Vert_F^2\leq \Vert A - A_d\Vert_F^2 +\tau\Vert A\Vert_F^2\,,
\end{align*}
where $A_d$ denotes the best rank-$d$ approximation to $A$ in the Frobenius sense. The rank $d$ should be selected to trade off approximation of $A$ and speed-ups for the method, e.g. such that $d/\tau\ll m+n$. We fall back on this approach to obtain a low-rank factorization of a distance matrix in linear time whenever needed, aware that this incurs an additional approximation (see Appendix~\ref{sec-lr-cost-mat}).

%\todo{add error due to the approximation of the cost?}

% In the following we introduce another way of using low-rank approximations by directly regularizing the nonnegative rank of the coupling involved in the GW problem.

%% file: sections/methods2.tex
\section{Low-rank Constraints for Couplings}
\label{sec:imposing}
In this section, we shift our attention to a different opportunity for speed-ups, \textit{without} Assumption~\ref{assump-low-rank}: we consider the GW problem on couplings that are \textit{low-rank}, in  the sense that they are factorized using two low-rank couplings linked by a common marginal $g$ in $\Delta_r^*$, the \textit{interior} of $\Delta_r$ (all entries positive). Writing the set of couplings with a nonnegative rank smaller than $r$~\citep[\S 3.1]{scetbon2021lowrank}:
\begin{align*}
\Pi_{a,b}(r) \eqdef 
 \begin{aligned}[t]
  \Big\{
  &P\in\mathbb{R}_{+}^{n\times m},\exists g\in\Delta_r^* \text{~s.t.~}  P=Q \Diag(1/g)R^T,\\
 & Q\in\Pi_{a,g}, \text{ and  } R\in\Pi_{b,g} %\right\}
  \Big\}\;,
 \end{aligned}
 \end{align*}
% $$\Pi_{a,b}(r) \eqdef \{P\in\Pi_{a,b}, \rank_{+}(P)\leq r\}.$$

% Writing
% \begin{align*}
% \Pi_{a,g,b} \eqdef 
%  \begin{aligned}[t]
%   \Big\{
%   &P\in\mathbb{R}_{+}^{n\times m},  P=Q \Diag(1/g)R^T,\\
%  & Q\in\Pi_{a,g}, \text{ and  } R\in\Pi_{b,g} %\right\}
%   \Big\}\,
%  \end{aligned}
%  \end{align*}
 
% \begin{align*}
% \Pi_{a,g,b} \eqdef 
%  \begin{aligned}[t]
%   \Big\{P\in\mathbb{R}_{+}^{n\times m},  P=Q \Diag(1/g)R^T,~Q\in\Pi_{a,g}, \text{ and  } R\in\Pi_{b,g} %\right\}
%   \Big\}.
%  \end{aligned}
%  \end{align*}
we can define the low-rank GW problem, written $\text{GW-LR}^{(r)}((a, \Dista),(b, \Distb))$ as the solution of
\begin{align}
\label{eq-GW-LR-reformulated}
 \min_{(Q,R,g)\in\mathcal{C}(a,b,r)} \mathcal{Q}_{\Dista,\Distb}(Q\Diag(1/g)R^T)\,,
\end{align}
where $\mathcal{C}(a,b,r) \eqdef \mathcal{C}_1(a,b,r)\cap \mathcal{C}_2(r)$, with
\begin{align*}
\mathcal{C}_1(a,b,r) &\eqdef 
 \begin{aligned}[t]
  \Big\{
  &(Q,R,g)\in\mathbb{R}_{+}^{n\times r}\times\mathbb{R}_{+}^{m\times r}\times(\mathbb{R}_{+}^{*})^{r} \\
    & \text{ s.t. }  Q\mathbf{1}_r=a, R\mathbf{1}_r=b
  \Big\}\,,
 \end{aligned}\\
 \mathcal{C}_2(r) &\eqdef 
 \begin{aligned}[t]
  \Big\{
  &(Q,R,g)\in\mathbb{R}_{+}^{n\times r}\times\mathbb{R}_{+}^{m\times r}\times\mathbb{R}^r_{+} \\
    & \text{ s.t. }  Q^T\mathbf{1}_n=R^T\mathbf{1}_m=g %\right\}
  \Big\}.
 \end{aligned}
 \end{align*}
% \begin{align*}
% \mathcal{C}_1(a,b,r) \eqdef 
%   \Big\{(Q,R,g)\in\mathbb{R}_{+}^{n\times r}\times\mathbb{R}_{+}^{m\times r}\times(\mathbb{R}_{+}^{*})^{r}~\text{ s.t. }  Q\mathbf{1}_r=a, R\mathbf{1}_r=b
%   \Big\},\\
% \mathcal{C}_2(r) \eqdef 
%   \Big\{(Q,R,g)\in\mathbb{R}_{+}^{n\times r}\times\mathbb{R}_{+}^{m\times r}\times\mathbb{R}^r_{+}~\text{ s.t. }  Q^T\mathbf{1}_n=R^T\mathbf{1}_m=g %\right\}
%   \Big\}.
%  \end{align*}
% \begin{align*}
% \mathcal{C}_1(a,b,r,\alpha) \eqdef 
%  \begin{aligned}[t]
%   \Big\{(Q,R,g)\in\mathbb{R}_{+}^{n\times r}\times\mathbb{R}_{+}^{m\times r}\times(\mathbb{R}_{+}^{*})^{r}~\text{ s.t. }  Q\mathbf{1}_r=a, R\mathbf{1}_r=b,~g\geq \alpha
%   \Big\}\;.
%  \end{aligned}
%  \end{align*}

% \begin{align}
% \label{eq-GW-LR-ent}
%  \text{GW-LR}^{(r)}_{\varepsilon,\alpha}((a, \Dista),(b, \Distb)) \eqdef\min_{(Q,R,g)\in\mathcal{C}(a,b,r,\alpha)} \mathcal{Q}_{\Dista,\Distb}(Q\Diag(1/g)R^T) - \varepsilon H((Q,R,g))\;.
% \end{align}
% \footnote{We also propose a double regularization scheme where in addition of constraining the nonnegative rank of the coupling, we regularize the objective by adding an entropic term in $(Q,R,g)$.
% See Appendix~\ref{sec:entropic-reg} for more details.}
\textbf{Mirror Descent Scheme.} We propose to use a MD scheme with respect to the generalized \text{KL} divergence to solve \eqref{eq-GW-LR-reformulated}. If one chooses $(Q_0,R_0,g_0)\in\mathcal{C}(a,b,r)$ an initial point such that $Q_0>0$ and $R_0>0$, this results in,
\begin{align}
\label{eq-barycenter-GW-inner}  
 (Q_{k+1},R_{k+1},g_{k+1}) \eqdef \!\! \argmin_{\zeta \in\mathcal{C}(a,b,r)} \!\! \text{KL}(\zeta,K_k)\; ,
\end{align}
where the three matrices $K_k \eqdef (K_{k}^{(1)},K_{k}^{(2)},K_{k}^{(3)})$ are
\begin{align*}
K_{k}^{(1)} &\eqdef \exp(4\gamma \Dista P_k\Distb R_k\Diag(1/g_k) +\log(Q_k))\\
K_{k}^{(2)} &\eqdef \exp(4\gamma \Distb P^T_k\Dista Q_k \Diag(1/g_k)+\log(R_k))\\
K_{k}^{(3)} &\eqdef \exp(-4\gamma\omega_k/g_k^2 + \log(g_k))
\end{align*}
with $[\omega_k]_i \eqdef [Q_k^T\Dista P_k\Distb R_k]_{i,i}$ for all $i\in\{1,\dots,r\}$ and $\gamma>0$ is a step size. Solving~\eqref{eq-barycenter-GW-inner} can be done efficiently thanks to Dykstra’s Algorithm as proposed in~\citep{scetbon2021lowrank}. See Algo.~\ref{alg-MDGW-LR} and Appendix~\ref{sec-Dykstra-app}.

\textbf{Avoiding vanishing components.} As in $k$-means optimization, the algorithm above might run into cases in which entries of the histogram $g$ vanish to $0$. Following ~\citep{scetbon2021lowrank} we can avoid this by setting a lower bound $\alpha$ on the weight vector $g$, such that $g\geq \alpha$ coordinate-wise. Practically, we introduce truncated feasible sets $\mathcal{C}(a,b,r,\alpha) \eqdef \mathcal{C}_1(a,b,r,\alpha)\cap \mathcal{C}_2(r)$ where $\mathcal{C}_1(a,b,r,\alpha)\eqdef\mathcal{C}_1(a,b,r)\cap\{(Q,R,g)~|~g\geq \alpha\}$.

% let us denote $\text{GW-LR}^{(r)}_{\alpha}((a, \Dista),(b, \Distb))$ the following optimization problem
% \begin{align}
% \label{eq-GW-LR-alpha}
%  \min_{(Q,R,g)\in\mathcal{C}(a,b,r,\alpha)} \mathcal{Q}_{\Dista,\Distb}(Q\Diag(1/g)R^T)\; .
% \end{align}

\textbf{Initialization.} To initialize our algorithm, we adapt the \textit{first lower bound} of~\citep{memoli2011gromov} to the low-rank setting and prove the following Proposition (see appendix~\ref{sec-proofs} for proof). 
\begin{prop}
\label{prop:init}
Let us denote $\tilde{x}=\Dista^{\odot2}a\in\mathbb{R}^n$, $\tilde{y}=\Distb^{\odot2}b\in\mathbb{R}^m$ 
% where $f(x)\eqdef x^2$ is a point-wise operation 
and $\tilde{C}=(|\sqrt{\tilde{x}_i}-\sqrt{\tilde{y}_j}|^2)_{i,j}\in\mathbb{R}^{n\times m}$. Then for all $r\geq 1$ we have,
\normalfont 
$$\text{GW-LR}^{(r)}_{\alpha}((a, \Dista),(b, \Distb))\geq 
      \text{LOT}^{(r)}_{\alpha}(\tilde{C},a,b)
    %  \min_{(Q,R,g)\in\mathcal{C}(a,b,r,\alpha)} \langle \tilde{C},Q\Diag(1/g)R^T\rangle  - \varepsilon H((Q,R,g))
    , \text{where}$$
 $$\text{LOT}^{(r)}_{\alpha}(\tilde{C},a,b):=\min_{(Q,R,g)\in\mathcal{C}(a,b,r,\alpha)} \langle \tilde{C},Q\Diag(1/g)R^T\rangle \; .$$
\end{prop}
$\text{LOT}^{(r)}_{\alpha}(\tilde{C},a,b)$ can be solved with~\citep{scetbon2021lowrank}. The cost $\tilde{C}$ is the squared Euclidean distance between two families $\{\tilde{x}_1,\dots,\tilde{x}_n\}$ and $\{\tilde{y}_1,\dots,\tilde{y}_m\}$ in 1-D, which admits a trivial rank 2 factorization. We can therefore apply the linear-time version of their algorithm to compute the lower bound. Algo.~\ref{alg-MDGW-LR} summarizes this, where $\mathcal{D}(\cdot)$ denotes the operator extracting the diagonal of a square matrix. In practice we observe that such initialization outperforms trivial or random initializations (see Section~\ref{sec-experiments-main}).
% Note that for all $k\geq 0$, $(Q_{k},R_{k},g_{k})$ live in $(\mathbb{R}_{+}^{*})^{n\times r}\times (\mathbb{R}_{+}^{*})^{m\times r} \times (\mathbb{R}_{+}^{*})^{r}$, and therefore $\bm{\xi}_k$ is well defined and lives also in $(\mathbb{R}_{+}^{*})^{n\times r}\times (\mathbb{R}_{+}^{*})^{m\times r} \times (\mathbb{R}_{+}^{*})^{r}$.
% \begin{wrapfigure}{r}{0.\textwidth}
% \vskip-.4cm
% \begin{minipage}{0.5\textwidth}

% \end{minipage}\vskip-.2cm
% \end{wrapfigure}

\textbf{Computational Cost.} Our initialization requires $\tilde{x}$ and $\tilde{y}$, obtained in $\mathcal{O}(n^2+m^2)$ operations. Running ~\citep[Algo.3]{scetbon2021lowrank} with a squared Euclidean distances between two families in 1-D has cost $\mathcal{O}((n+m)r)$. Solving the barycenter problem as defined in~\eqref{eq-barycenter-GW-inner} can be done efficiently thanks to Dykstra’s Algorithm. Indeed, each iteration of~\citep[Algo. 2]{scetbon2021lowrank}, assuming $(K^{(1)}_k,K^{(2)}_k,K^{(3)}_k)$ is given, requires only $\mathcal{O}((n+m)r)$  algebraic operations. However, computing kernel matrices $(K^{(1)}_k,K^{(2)}_k,K^{(3)}_k)$ at each iteration of Algorithm~\ref{alg-MDGW-LR} requires a quadratic complexity with respect to the number of samples. Overall the proposed algorithm, while faster than the cubic implementation proposed in~\citep{peyre2016gromov}, still needs $\mathcal{O}((n^2+m^2)r)$ operations per iteration. 

\textbf{Dykstra Iterations.}
In our complexity analysis, we do not take into account the number of iterations required to terminate Dykstra's Algorithm. We show experimentally~(see Fig.~\ref{fig-iteration}) that, as usually observed for Sinkhorn~\citep[Fig. 5]{cuturi2013sinkhorn}, this number does not depend on problem size $n,m$, but rather on the geometric characteristics of $\Dista, \Distb$ and $\gamma$.

%In the following we will see that by combining both nonnegative low-rank constraints on the coupling and low-rank approximations of the distance matrices, we can obtain a linear time algorithm with respect to the number of samples which computes an approximation of the GW distance.

\textbf{Convergence of MD.} Although objective~(\ref{eq-GW-LR-reformulated}) is not convex in $(Q,R,g)$, we obtain the non-asymptotic stationary convergence of our proposed method. In~\citep{scetbon2021lowrank}, the authors study the convergence of the MD scheme when applied to the low-rank formulation of OT. In the GW setting, such strategy makes even more sense as the GW problem is a NP-hard non-convex problem and obtaining global guarantees is out of reach in a general framework. Therefore we follow the strategy proposed in~\citep{scetbon2021lowrank} and consider the following convergence criterion,
\begin{align*}
   \Delta_{\alpha}(\xi,\gamma) \eqdef \frac{1}{\gamma^2}(\text{KL}(\xi,\mathcal{G}_{\alpha}(\xi,\gamma))+\text{KL}(\mathcal{G}_{\alpha}(\xi,\gamma),\xi))
\end{align*}
where $\mathcal{G}_{\alpha}(\xi,\gamma) \eqdef \argmin_{\zeta\in \mathcal{C}(a,b,r,\alpha)}\{ \langle \nabla \mathcal{Q}_{\Dista,\Distb}(\xi),\zeta \rangle +\frac{1}{\gamma} \text{KL}(\zeta,\xi) \}.$ This convergence criterion is in fact stronger than the one using the (generalized) projected gradient presented in~\cite{ghadimi2013minibatch} to obtain non-asymptotic stationary convergence of the MD scheme. Indeed the criterion used there is defined as the square norm of the following vector:
\begin{align*}
   P_{\mathcal{C}(a,b,r,\alpha)}(\xi,\gamma) \eqdef \frac{1}{\gamma}(\xi-\mathcal{G}_{\alpha}(\xi,\gamma))\; ,
\end{align*}
which can be seen as a generalized projected gradient of $\mathcal{Q}_{\Dista,\Distb}$ at $\xi$. By denoting $X \eqdef \mathbb{R}^d$ and by replacing the \emph{Bregman Divergence} $\text{KL}(\zeta,\xi)$ by $\frac{1}{2}\Vert \zeta-\xi\Vert_2^2$ in the MD scheme, we would have $P_{X}(\xi,\gamma)=\nabla\mathcal{Q}_{\Dista,\Distb}(\xi)$. Now observe that we have
% Here we consider instead the following criterion to establish convergence:
% \begin{align*}
%   \Delta_{\varepsilon,\alpha}(\bm{\xi},\gamma) \eqdef \frac{1}{\gamma^2}(\mathrm{KL}(\bm{\xi},\mathcal{G}_{\varepsilon,\alpha}(\bm{\xi},\gamma))+\mathrm{KL}(\mathcal{G}_{\varepsilon,\alpha}(\bm{\xi},\gamma),\bm{\xi})).
% \end{align*}
% Such criterion is in fact stronger than the one used in~\cite{ghadimi2013minibatch} as we have
\begin{align*}
    \Delta_{\alpha}(\xi,\gamma)&=\frac{1}{\gamma^2}(\langle  \nabla h(\mathcal{G}_{\alpha}(\xi,\gamma)) - \nabla h(\xi),  \mathcal{G}_{\alpha}(\xi,\gamma) - \xi\rangle\\
    &\geq \frac{1}{2\gamma^2}  \Vert \mathcal{G}_{\alpha}(\xi,\gamma) - \xi\Vert_1^2\\
    &=\frac{1}{2}\Vert  P_{\mathcal{C}(a,b,r,\alpha)}(\xi,\gamma)\Vert_1^2
\end{align*}
where $h$ denotes the minus entropy function and the last inequality comes from the strong convexity of $h$ on $\mathcal{C}(a,b,r,\alpha)$. Therefore  $\Delta_{\alpha}(\xi,\gamma)$ dominates $\Vert  P_{\mathcal{C}(a,b,r,\alpha)}(\xi,\gamma)\Vert_1$ and characterizes a stronger convergence.

For any $1/r\geq \alpha> 0$, Proposition~\ref{prop:cvg-MD-Dykstra} shows the non-asymptotic stationary convergence of the MD scheme for Problem~\eqref{eq-GW-LR-reformulated}. See Appendix~\ref{sec-proofs} for the proof.
\begin{prop}
\label{prop:cvg-MD-Dykstra}
Let $\frac{1}{r}\geq \alpha> 0, N\geq 1$ and $L_{\alpha} \eqdef 27(\Vert \Dista\Vert_2\Vert \Distb\Vert_2/\alpha^4)$. Consider a constant stepsize $\gamma=\frac{1}{2L_{\alpha}}$ in the MD scheme~(\ref{eq-barycenter-GW-inner}). Writing $D_0 \eqdef  \mathcal{Q}_{\Dista,\Distb}(Q_0\Diag(1/g_0)R_0^T)- \text{\normalfont{GW-LR}}^{(r)}_{\alpha}((a, \Dista),(b, \Distb))$ the gap between initial value and optimum, one has
\begin{align*}
    \min_{1\leq k\leq N}\Delta_{\alpha}((Q_k,R_k,g_k),\gamma)\leq \frac{4L_{\alpha}  D_0}{N}.
\end{align*}
\end{prop}
Since for $\alpha$ small enough, $\text{GW-LR}^{(r)}_{\alpha}((a, \Dista),(b, \Distb))=\text{GW-LR}^{(r)}((a, \Dista),(b, \Distb))$, Proposition~\ref{prop:cvg-MD-Dykstra} shows that our algorithm reaches a stationary point of~(\ref{eq-GW-LR-reformulated}).

This Proposition claims that within at most $N$ iterations the minimum of the $(\Delta_{\alpha}((Q_t,R_t,g_t),\gamma))_{1\leq t\leq N}$ is of order $\mathcal{O}(1/N)$. Note that this is a standard way to obtain the stationary convergence (see e.g.~\cite{ghadimi2013minibatch}. In practice, this is sufficient to define a stopping criteria, as one could simply compute at each iteration the criterion and keep only in memory the smallest value at each iteration.

\begin{algorithm}
\SetAlgoLined
\textbf{Inputs:} $\Dista,\Distb,a,b,r,\alpha,\gamma$\\
$\tilde{x}\gets \Dista^{\odot2}a$, $\tilde{y}\gets \Distb^{\odot2}b$  \,\,\,\,\,{\color{red}{$\mathtt{m^2+n^2}$}} \label{line-step-1} \\
$z_1\gets \tilde{x}^{\odot 2}$, $z_2\gets \tilde{y}^{\odot 2}$ \,\,\,\,\,{\color{mygreen}{$\mathtt{m+n}$}}\\
$\tilde{C}_1\gets [z_1,\mathbf{1}_n,-
\sqrt{2}\tilde{x}]$, $\tilde{C}_2\gets [\mathbf{1}_m,z_2,\sqrt{2}\tilde{y}]^T$ \,\,\,\,\,{\color{mygreen}{$\mathtt{n+m}$}}\\
$(Q,R,g)\gets \text{LOT}^{(r)}_{\alpha}(\tilde{C}_1\tilde{C}_2,a,b)$ \,\,\,\,\,{\color{mygreen}{$\mathcal{O}(\mathtt{(n+m)r})$}}\\
\For{$t=1,\dots$}{
    $C_1\gets -\Dista Q\Diag(1/g)$\,\,\,\,\,{\color{red}{$\mathcal{O}(\mathtt{n^2r})$}}\label{line-step-2}\\
    $C_2\gets R^T\Distb$\,\,\,\,\,{\color{red}{$\mathcal{O}(\mathtt{m^2r})$}}\label{line-step-3}\\
    $K^{(1)}\gets Q\odot e^{4\gamma C_1C_2 R \Diag(1/g)}$\,\,{\color{mygreen}{$\mathcal{O}(\mathtt{(m+n)r^2})$}}\\ 
    $K^{(2)}\gets R\odot e^{4\gamma C_2^TC_1^TQ \Diag(1/g)}$\,\,{\color{mygreen}{$\mathcal{O}(\mathtt{(m+n)r^2})$}}\\
    $\omega\gets \mathcal{D}(Q^TC_1C_2R)$\,\,{\color{mygreen}{$\mathcal{O}(\mathtt{nr^2})$}}\\
    $K^{(3)}\gets g\odot e^{-4\gamma\omega/g^2}$\,\,{\color{mygreen}{$\mathcal{O}(\mathtt{r})$}}\\
    $Q,R,g\gets \argmin\limits_{\zeta \in\mathcal{C}(a,b,r,\alpha)} \!\! \text{KL}(\zeta,K)$\,\,{\color{mygreen}{$\mathcal{O}(\mathtt{(m+n)r})$}}
  }
   $c_1\gets \langle\tilde{x},a\rangle +  \langle\tilde{y},b\rangle$\,\,\,\,\,{\color{mygreen}{$\mathtt{n+m}$}}\\
 $C_1\gets -\Dista Q\Diag(1/g)$\,\,\,\,\,{\color{red}{$\mathcal{O}(\mathtt{n^2r})$}}\label{line-step-4}\\
 $C_2\gets R^T\Distb$\,\,\,\,\,{\color{red}{$\mathcal{O}(\mathtt{m^2r})$}}\label{line-step-5}\\
 $G\gets C_2R$,  $G\gets C_1 G$\,\,\,\,\,{\color{mygreen}{$\mathcal{O}(\mathtt{(m+n)r^2})$}}\\
 $c_2\gets - 2 \langle Q,G \Diag(1/g)\rangle $\,\,\,\,\,{\color{mygreen}{$\mathcal{O}(\mathtt{nr})$}}\\
 $\mathcal{Q}\gets c_1 + c_2$ \\
\textbf{Return:} $\mathcal{Q}$
\caption{Low-Rank GW \label{alg-MDGW-LR}
% , $\text{GW-LR}_{\varepsilon,\alpha}^{(r)}((a,\Dista), (b,\Distb))$ \label{alg-MDGW-LR}
}
\end{algorithm}

%% file: sections/methods3.tex
\section{Double Low-rank GW}
\label{sec-lin-GW}
Almost all operations in Algorithm~\ref{alg-MDGW-LR} only require linear memory storage and time, except for the computations of $\tilde{x}=\Dista^{\odot2}a$  and $\tilde{y}=\Distb^{\odot2}b$ in Line~\ref{line-step-1}, and the four updates involving $C_1$ and $C_2$ in Lines~\ref{line-step-2},\ref{line-step-3},\ref{line-step-4},\ref{line-step-5}
which all require a quadratic number of algebraic operations. When adding Assumption~\ref{assump-low-rank} from \S\ref{sec-LR-cost} to the rank constrained approach from \S\ref{sec:imposing}, we show that the strengths of both approaches can work hand in hand, both in easier initial evaluations of $\tilde{x}, \tilde{y}$, but, most importantly, at each new recomputation of a \textit{factorized} linearization of the quadratic objective: 

\textbf{Linear-time Norms in Line~\ref{line-step-1}} Because $\Dista$ admits a low-rank factorization, one can obtain a low-rank factorization for $\Dista^{\odot2}$ pending the condition $d^2\ll n$. Indeed, remark that for $u,v\in\mathbb{R}^d, \langle u,v\rangle^2=\langle u u^T, v v^T\rangle$. Therefore, if one describes $\facA:=[u_1;\dots;u_n]$ and $\facAA:=[v_1;\dots;v_n]$ row-wise, and one uses the flattened out-product operator $\psi(u):= \text{vec}(uu^T)\in\mathbb{R}^{d^2}$ where $\text{vec}(\cdot)$ flattens a matrix,
\begin{align*}
    \Dista^{\odot2}=\tilde{\facA}\tilde{\facAA}^T  \text{ where } \tilde{\facA} & = [\psi(u_1;\dots;\psi(u_n)],\; \\
    \tilde{\facAA} & = [\psi(v_1);\dots;\psi(v_n)]\;.
\end{align*}
Line~\ref{line-step-1} in Algo.~\ref{alg-MDGW-LR} can be replaced by 
$\tilde{x} \gets \tilde{\facA}\tilde{\facAA}^T a$
and
$\tilde{y} \gets \tilde{\facB}\tilde{\facBB}^Tb$.
Pending the condition $d^2\ll n, d'^2\ll m$, this results in $nd^2 + m(d')^2$ operations. Note that Algo.~\ref{alg-quad-GW} (line \ref{line-step-9-alg-2}) can also benefit from this factorization, however as its complexity is already quadratic, the linearization of this operation has no effect on the global computational cost.
%Finally computing $\Dista^2a=\tilde{\facA}\tilde{\facAA}^Ta$ and $(\Distb)^2a=\tilde{\facB}\tilde{\facBB}^Tb$ can be done in respectively $\mathcal{O}(nd^2)$ and $\mathcal{O}(m(d')^2)$ operations. 

\textbf{Linearization of Lines~\ref{line-step-2},\ref{line-step-3},\ref{line-step-4},\ref{line-step-5}.} The critical step in Algo.~\ref{alg-mirror-descent} that requires updating $C$ at each outer iteration is cubic. As described earlier in Algo.~\ref{alg-MDGW-LR} and Algo.~\ref{alg-quad-GW}, a low-rank constraint on the coupling or a low-rank assumption on costs $\Dista$ and $\Distb$ reduce this cost to quadratic. Remarkably, both can be combined to yield linear time by replacing in Algo.~\ref{alg-MDGW-LR}, Lines~\ref{line-step-2}, \ref{line-step-3}, \ref{line-step-4},  \ref{line-step-5} by 
\begin{align*}
 C_1 \leftarrow  -\facA\facAA^TQ\Diag(1/g)
 \quad\text{and}\quad 
 C_2 \leftarrow R^T\facBB\facB^T\;.
\end{align*}
% though the identities:
%\begin{align*}
% C_1 = -\Dista Q\Diag(1/g) = -\facA\facAA^TQ\Diag(1/g)~~\text{and } C_2 = R^T\Distb = R^T\facBB\facB^T\;.
%\end{align*}
Note that this speed-up would not be achieved using other approaches that output a low rank approximation of the transport plan~\citep{altschuler2018approximating,altschuler2018massively,scetbon2020linear}. The crucial obstacle to using these methods here is that the cost matrix $C$ in GW changes throughout iterations, and is synthetic--the output of a matrix product $\Dista P\Distb$ involving the very last transport $P$. This stands in stark contrast with the requirements in ~\citep{altschuler2018approximating,altschuler2018massively,scetbon2020linear} that the \textit{kernel} matrix corresponding to $K_\varepsilon=e^{-C/\varepsilon}$ admits favorable properties, such as being p.s.d or admitting an explicit (random or not) finite dimensional feature approximation.

% Combining the results in \S\ref{sec:imposing} with those from \S\ref{sec-LR-cost} results in updates for $C_1$ and $C_2$ that only require $\mathcal{O}(nrd)$ and $\mathcal{O}(mrd')$ operations.

\textbf{Linear time GW.} We have shown that (red) quadratic operations appearing in Algo.~\eqref{alg-MDGW-LR} can be replaced by linear alternatives. The iterations that have not been modified had an overall complexity of $\mathcal{O}(mr(r+d') + nr(r + d))$. The initialization and linearization steps can now be performed in linear time and complexity,  respectively in $\mathcal{O}(n(r+d^2) + m((d')^2+r))$ and  $\mathcal{O}((nr(r+d) +mr (r+d'))$.

%% file: sections/experiment.tex
\section{Experiments}
\label{sec-experiments-main}
Our goal in this section is to provide practical guidance on how to use our method (to set stepsize $\gamma$, lower bound $\alpha$ on entries of $g$ and rank $r$) and compare its practical performance with other baselines, both in terms of running times and relevance, on 5 simulated datasets and 2 real world applications. We consider our quadratic approach \textbf{LR} (Algo.~\ref{alg-MDGW-LR}) and its linear time counterpart \textbf{Lin LR} (\S\ref{sec-lin-GW}). We compare them with \textbf{Ent}, the cubic implementation of~\citep{peyre2016gromov}, and its improved quadratic version \textbf{Quad Ent} introduced in this paper (Algo.~\ref{alg-quad-GW}). We also use \textbf{MREC} as implemented in~\cite{blumberg2020mrec}. 
Because all these approaches admit different hyperparameters, we evaluate them by stressing GW loss as a function of computational effort, as well as performance in downstream metrics. Because the couplings obtained by \textbf{MREC} do \textit{not} satisfy marginal constraints, computing its GW loss is irrelevant, but its matching can be used in the single cell genomics experiments we consider. Experiments were run on a MacBook Pro 2019 laptop, and data from \texttt{github.com/rsinghlab/SCOT}. The code is available at \href{https://github.com/meyerscetbon/LinearGromov}{https://github.com/meyerscetbon/LinearGromov}.

\begin{figure}[!t]
\centering
\includegraphics[width=0.49\textwidth]{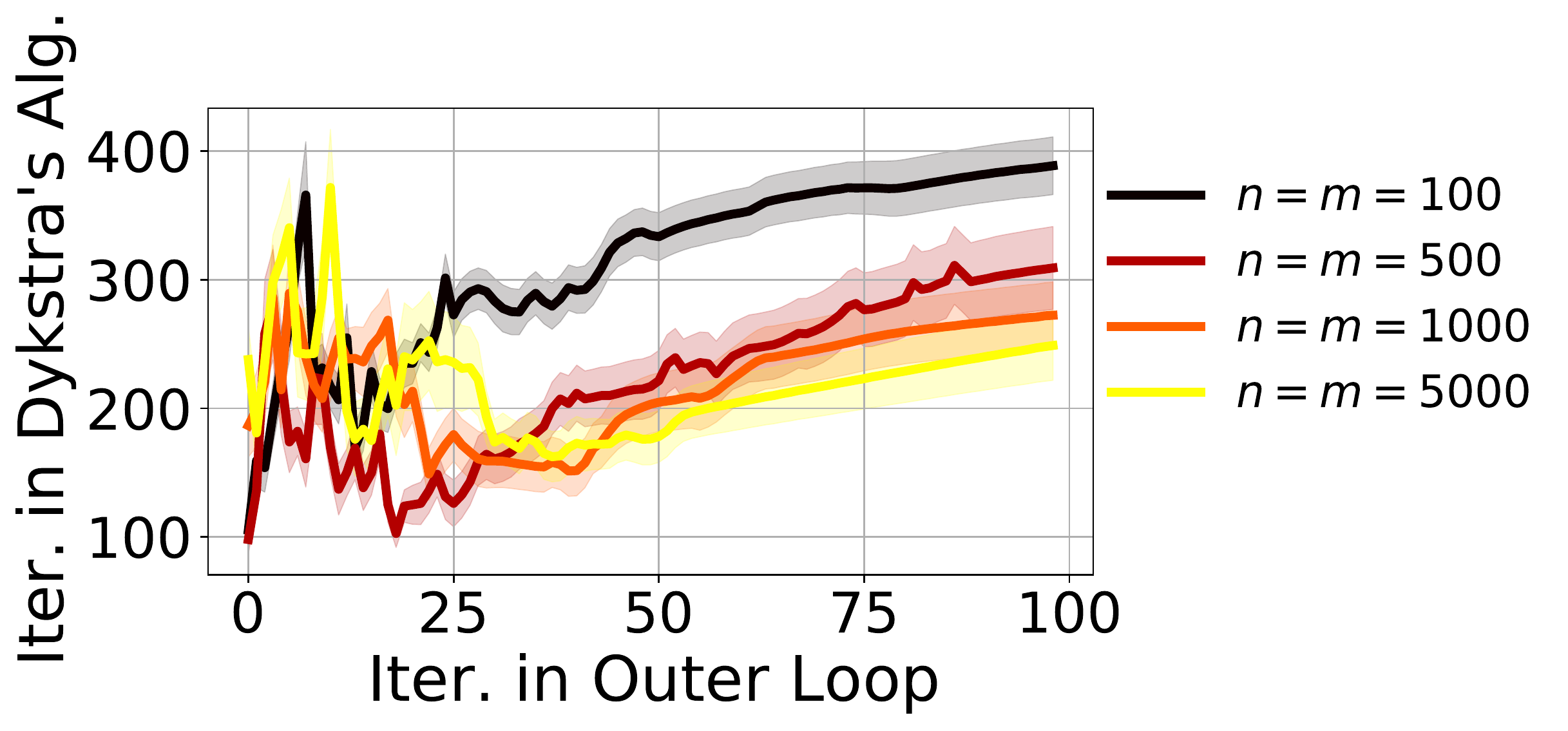}
\caption{We consider samples of a mixture of 10 anisotropic Gaussians in resp. 10 and 15-D endowed with the squared Eucl. metric. The number of iterations of Dykstra's algorithm required to reach a precision of $\delta=1e-3$ along the iterations of the Algo.~\ref{alg-MDGW-LR} is not impacted significantly by varying $n$, the sample size.
 \label{fig-iteration}}
% \vskip-.4cm
\end{figure}

\textbf{Initialization.} For a fair comparison with the entropic approach, we adapt the \textit{first lower bound} of~\citep[Def. 6.1]{memoli2011gromov} to the entropic case to initialize it.
% to initialize them all.both \textbf{Entropic-GW} and \textbf{Quad Entropic-GW}. 
In all experiments displaying time-accuracy tradeoffs, we report computation budget as number of operations. Accuracy is measured by evaluating the ground-truth energy $\mathcal{Q}_{\Dista,\Distb}$ (even in scenarios when the method uses a low rank approximation for $\Dista, \Distb$ at optimization time). We repeat all experiments 10 times on random resampling of the measures in all synthetic problems, to obtain error bars.

\begin{figure*}[t!]
    \centering
  \includegraphics[width=1\textwidth]{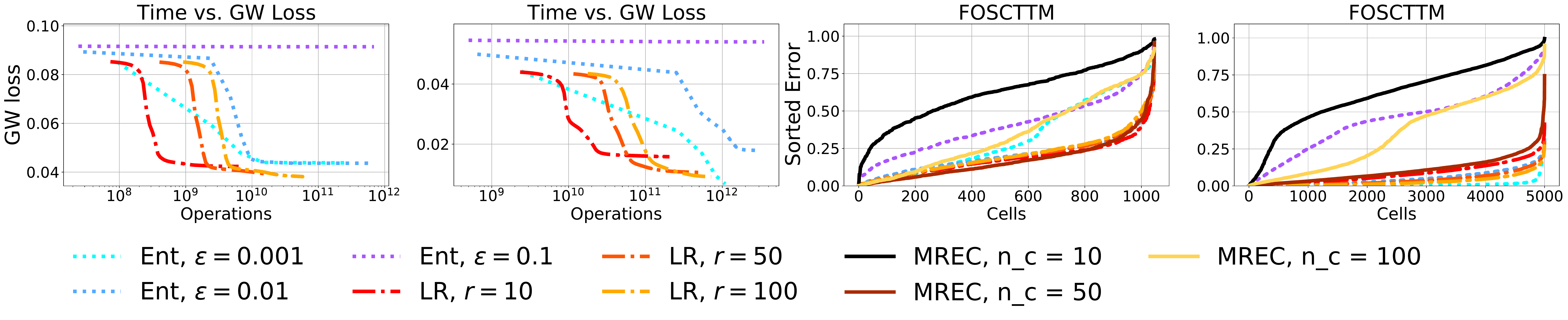}
\caption{We consider both the SNAREseq dataset (\emph{left, middle-right}) which consists in two point clouds of $n=m=1047$ samples in respectively 10-D and 19-D and the Splatter dataset (\emph{middle-left, right}) composed of two point clouds of $n=m=5000$ samples in respectively 50-D and 500-D. The cost considered is the shortest-path distance of a $k-NN$ graph. We compare both the time-accuracy tradeoffs of our method with the Entropic-GW (\emph{left, middle-left}) and the FOSCTTMs ranked in the increasing order of \textbf{LR}, \textbf{Ent} and \textbf{MREC} when varying their hyperparameters (\emph{middle-right, right}). Because the coupling returned by \textbf{MREC} does not satisfy marginal constraints, we do not include it in left plots. Our method reaches similar accuracy while being order of magnitude faster than \textbf{Ent} even for a small rank $r=n/100$. We notice that the alignments obtained by our method are robust to the choice of $r$, with similar performance for all methods.}\label{fig-scot}
\vspace{-0.4cm}
\end{figure*}

\textbf{On the iterations of Dykstra's Algorithm.} In this experiment, we show that the number of iterations involved in the Dykstra's Algorithm does not depends on $n$ the number of samples when applying Algo.~\ref{alg-MDGW-LR}. In Fig.~\ref{fig-iteration}, we consider samples of mixtures of (10 and 15) anisotropic Gaussians in resp. 10 and 15-D and report the number of iterations of the Dykstra's Algorithm required to reach a precision $\delta=1e-3$ along the iterations of Algo.~\ref{alg-MDGW-LR}. We observe that the number of iterations in Dykstra does not depend on $n$ the number of samples considered. Note that for all the sample sizes considered, we need far fewer iterations (usually $\leq 25$) for the outer loop to converge: the plots show a larger $x$-axis than what is observed in practice.

% \begin{wrapfigure}{r}{0.45\textwidth}
% \begin{minipage}{0.45\textwidth}

% \begin{figure*}[!t]
% \centering
% \includegraphics[width=1\textwidth]{figures/Bloc_E/plot_accuracy_vs_time_2.png}
% \caption{The number of cluster in each distribution is 10 and the number of samples is $n=m=5000$. The ground cost is the Euclidean distance. As we can evaluate the distance between two arbitrary points, we can obtain in linear-time an efficient approximation of the distance matrices $\Dista$ and $\Distb$ as presented in~\ref{sec-LR-cost}. The rank of their factorizations is fixed to be $d=d'=100$. \textbf{GW-LR} and \textbf{Entropic-GW} corresponds to the case where the full matrices $\Dista$ and $\Distb$ are considered while \textbf{Lin GW-LR} and \textbf{Quad Entropic-GW} take as inputs the low-rank approximations of the distance matrices. We plot the time-accuracy tradeoff for multiple choices of $\gamma$ and rank $r$ defined as a fraction of $n$. For \textbf{Entropic-GW} and \textbf{Quad Entropic-GW}, we set $\varepsilon=1/\gamma$ as proposed in~\cite{peyre2016gromov}. Recall that for low-rank methods, we set $\varepsilon=0$.}\label{fig-bloc-Euclidean}
% \vspace{-0.1cm}
% \end{figure*}
% For \textbf{GW-LR} and \textbf{Lin GW-LR}, and in all experiments, we set the lower bound on entries of $g$ to $\alpha=10^{-10}$. 

\paragraph{Sensitity to $\gamma$ and $\alpha$.} 
% Recall that our method depends on three hyperparameters which are the lower-bound $\alpha$ on the common marginal $g$, the step-size $\gamma$ of the MD scheme and the rank $r$ imposed on the couplings. 
We study how optimization parameters $\gamma$ and $\alpha$ impact results. We consider $n=m=1000$ samples drawn from two mixtures of (2 and 3) anisotropic Gaussians in respectively 5-D and 10-D (details in Appendix~\ref{sec-hyperparam}). Fig.~\ref{fig-gamma}, reports the time vs. GW loss tradeoff of our method when varying $\gamma$, both for $r=n/100$ or $n/10$ illustrating its robustness to that choice. Fig.~\ref{fig-alpha} in Appendix~\ref{sec-hyperparam} shows similar conclusions with respect to $\alpha$. Recall that $\alpha$ was only used to lower bound the weights of barycenter $g$, to ensure no collapse. In all other experiments, we always set $\gamma=100$ and $\alpha=10^{-10}$ for our methods, and only focus on rank $r$.

\begin{wrapfigure}{r}{.26\textwidth}
% \vskip-.3cm
\includegraphics[width=0.25\textwidth]{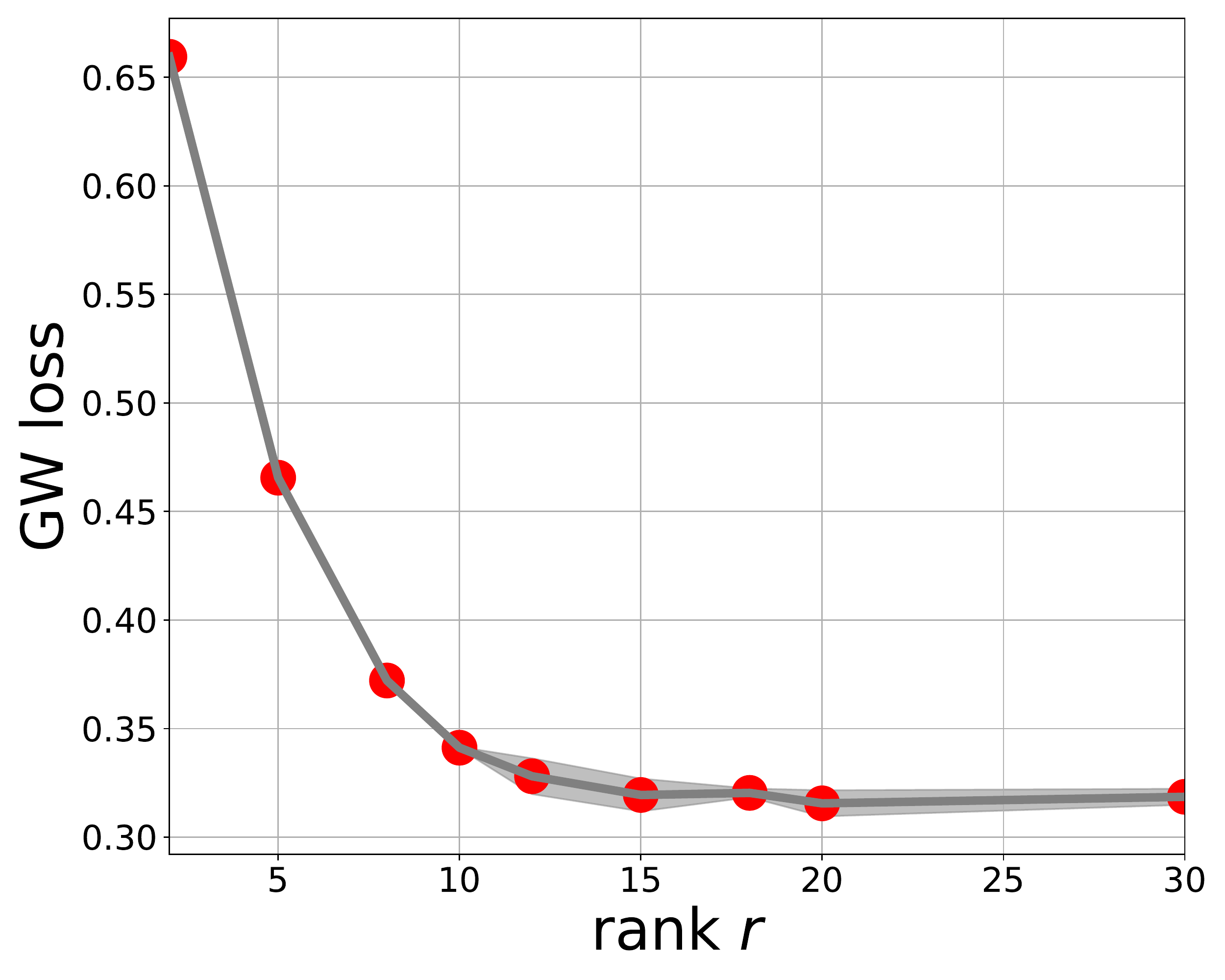}
% \vskip-.3cm
\end{wrapfigure}
\paragraph{Effect of the rank.}  We study the impact of rank $r$ on our method. We consider samples from two Gaussian mixtures, with respectively 10 and 20 centers in 10-D and 15-D and $n=m=5000$. We compute the GW cost obtained by \textbf{Lin LR} in the squared Euclidean setting as a function of $r$ the rank. We observe that the loss decreases as the rank increases until the rank $r$ reaches 20 (the largest number of clusters in our mixtures). Therefore, our method is able to capture the clustered structure of data (See Appendix~\ref{sec-effect-rank}). In practice $r$ should be selected such that it corresponds to the number of clusters in the data.

\begin{figure}[!t]
    \centering
   \includegraphics[width=0.48\textwidth]{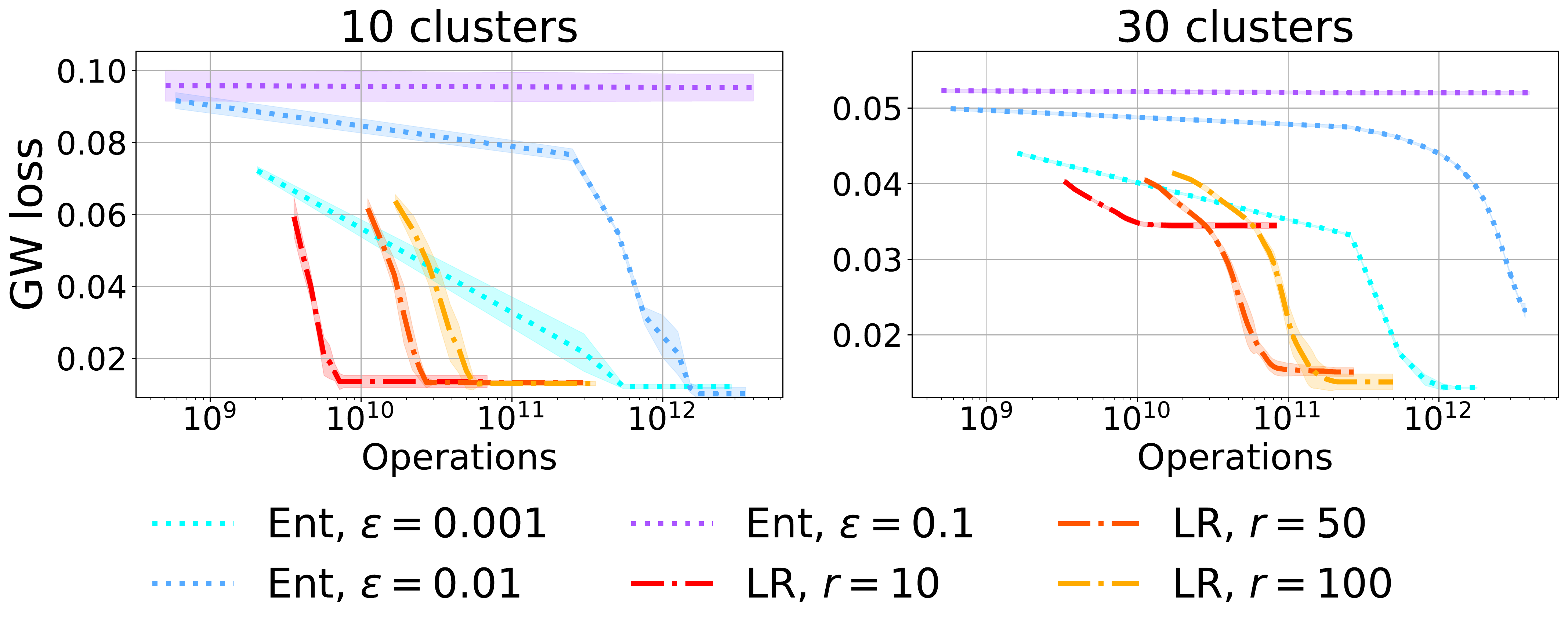}
  \caption{We sample $n=m=5000$ points from two anisotropic Gaussian blobs, respectively in 10 and 15-D, with either 10 or 30 clusters, endowed with the Euclidean distance. We compare our quadratic method \textbf{LR} with the cubic Entropic GW \textbf{Ent}, which requires instantiating matrices $\Dista$ and $\Distb$. We vary both $r$ (our method) and $\varepsilon$ (entropic). Our method obtains similar GW loss, while being orders of magnitude faster. Note the gap in performance between $r=10$ and $r=50$ when the input measures have 30 clusters: the GW loss decreases as the rank $r$ increases until it reaches the number of clusters in the data.}\label{fig-bloc-Euclidean}
\vspace{-.2cm}
\end{figure}

\begin{figure}
    \centering
   \includegraphics[width=0.48\textwidth]{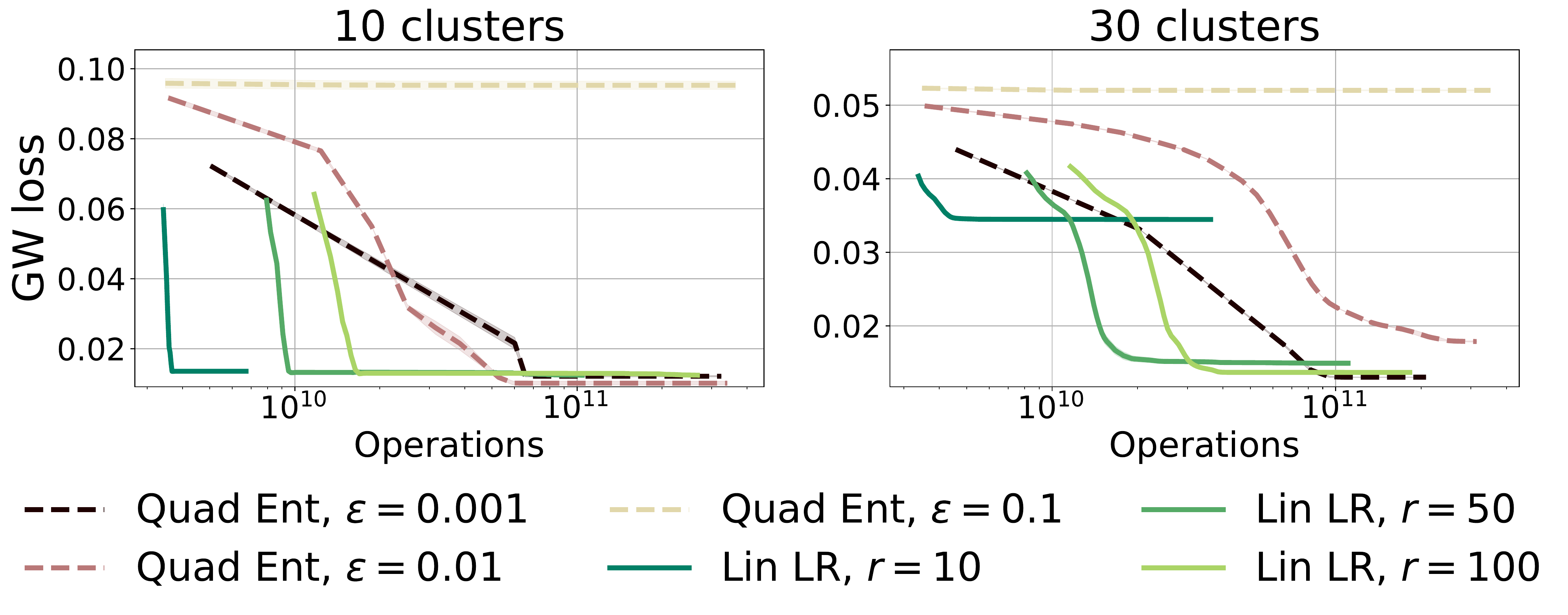}
\caption{Same setting as Fig.~\ref{fig-bloc-Euclidean}, using a low-rank approximation of the Euclidean distance (see~\S\ref{sec-LR-cost}) to introduce our linear method \textbf{Lin LR} and compare it with \textbf{Quad Ent}. The rank of their factorizations is set to $d=d'=100$. We vary $\varepsilon$ and rank $r$ to reach similar conclusions to those outlined in Fig.~\ref{fig-bloc-Euclidean}. Note also that both \textbf{Lin LR} and \textbf{Quad Ent} reach similar GW loss as those obtained by their full-rank counterparts, while being orders of magnitude faster.}\label{fig-bloc-Euclidean-factorized}
\vspace{-.1cm}
\end{figure}

\begin{figure}[!t]
    \centering
  \includegraphics[width=0.48\textwidth]{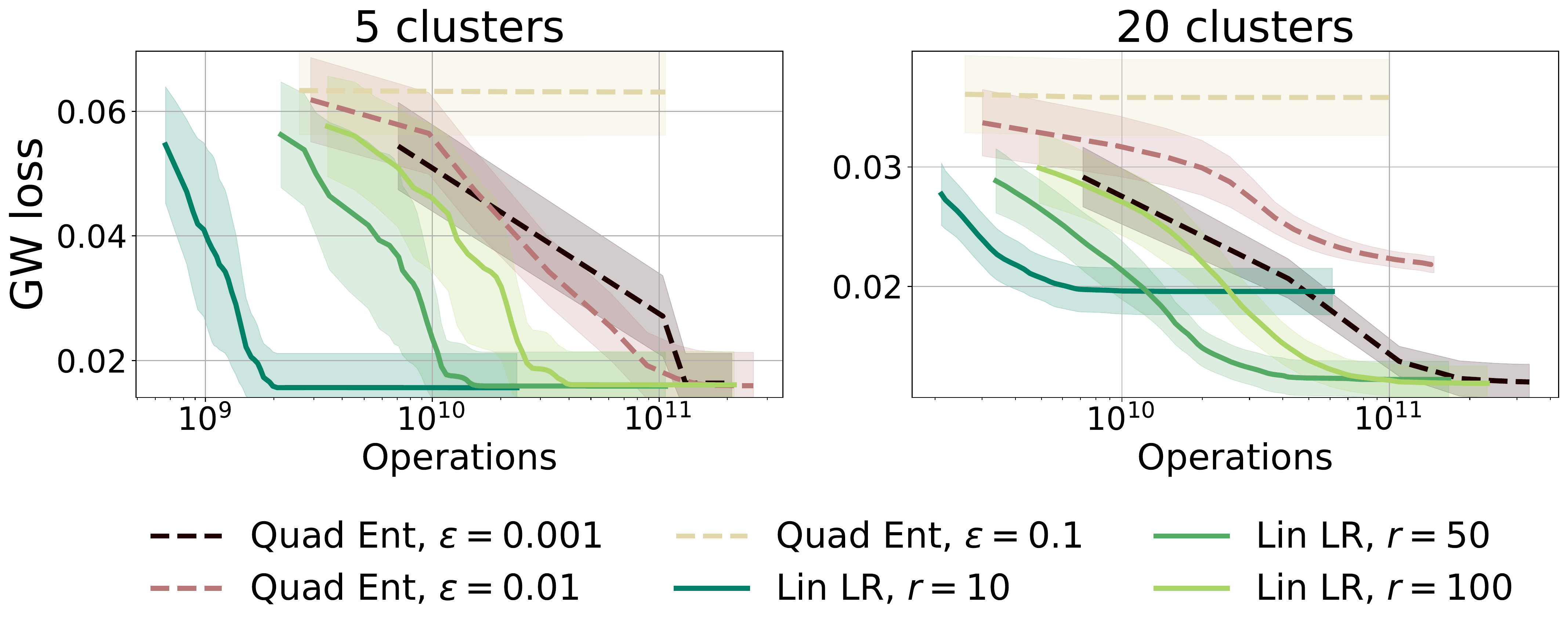}
\caption{Setting as in Fig.~\ref{fig-bloc-Euclidean}, with $n=m=10000$ samples from anisotropic Gaussian blobs of 5 or 20 clusters, endowed with the squared Eucl. distance. We compare \textbf{Lin LR} and \textbf{Quad Ent} using exact factorizations of $\Dista$ and $\Distb$.}\label{fig-bloc-SEuclidean}
\vspace{-0.1cm}
\end{figure}

\paragraph{Synthetic low-rank problem.} We consider two anisotropic Gaussian blobs with the same number of blobs in respectively 10-D and 15-D. We constrain the distance between the centroids of the clusters to be larger than the dimension (see Appendix~\ref{sec-low-rank-problem} for illustrations). In Figures~\ref{fig-bloc-Euclidean} and \ref{fig-bloc-Euclidean-factorized}, when the underlying cost is the (\textit{not} squared) Euclidean distance, our methods manage to consistently obtain similar GW loss that those obtained by entropic methods, using very low rank $r=n/100$, while being orders of magnitude faster. Fig.~\ref{fig-bloc-SEuclidean} explores the more favorable case where the underlying cost is the \textit{squared} Euclidean distance, reaching similar conclusions. 

\begin{figure}
% \vspace{-0.4cm}
\centering
\includegraphics[width=0.48\textwidth]{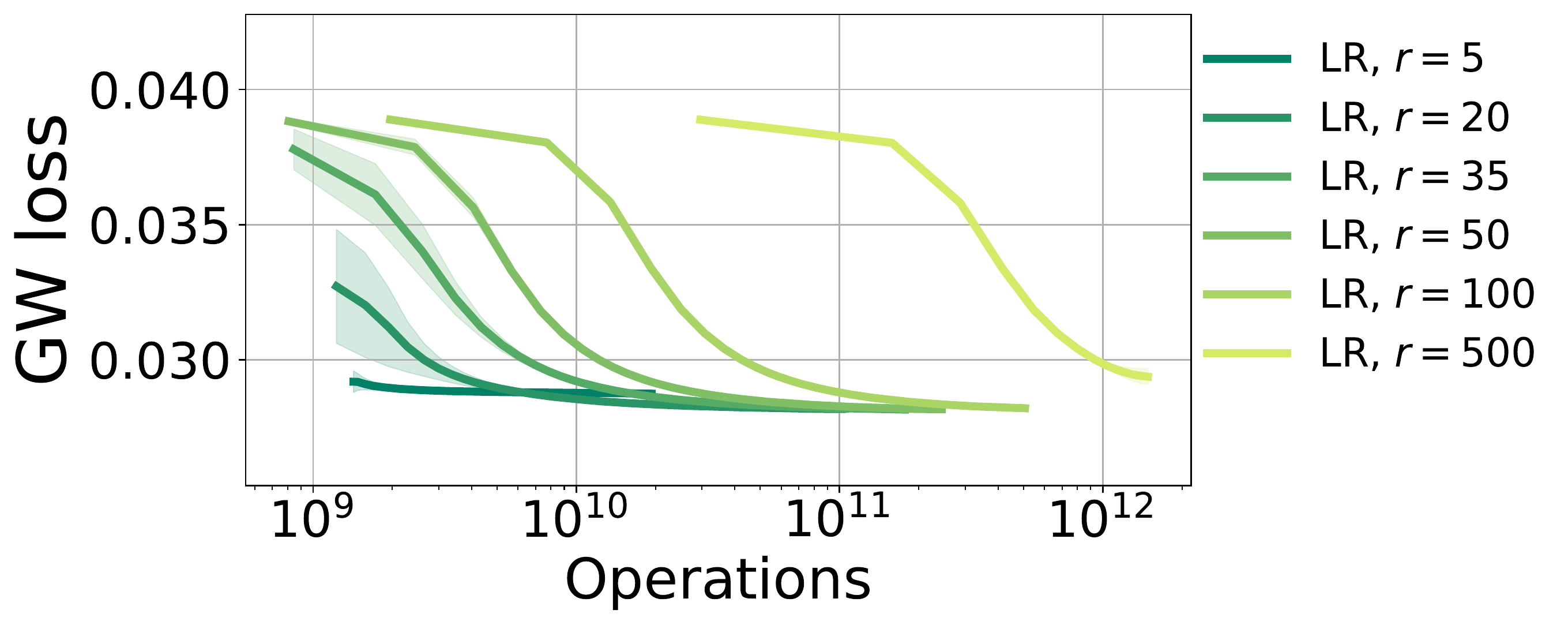}
\caption{We sample $n=1e5$ points from the unit square in 2-D. The underlying cost considered is the squared Euclidean cost. In this regime, only \textbf{Lin GW-LR} can be computed. We plot the time/loss tradeoff when varying $r$.
}\label{fig-large-scale}
% \vspace{-0.1cm}
\end{figure}

\paragraph{Large scale experiment.} In this experiment, we show that our method is able to compute an approximation of the GW cost in the large sample setting. In Fig.~\ref{fig-large-scale}, we samples $n=m=1e5$ samples from the unit square in 2-D and we compare the time/loss tradeoff when varying the rank $r$. We show that our method is the only one able to approach the GW cost in such regimes.

% \begin{wrapfigure}{l}{0.5\textwidth}
\begin{figure}[!h]
\centering
\includegraphics[width=0.48\textwidth]{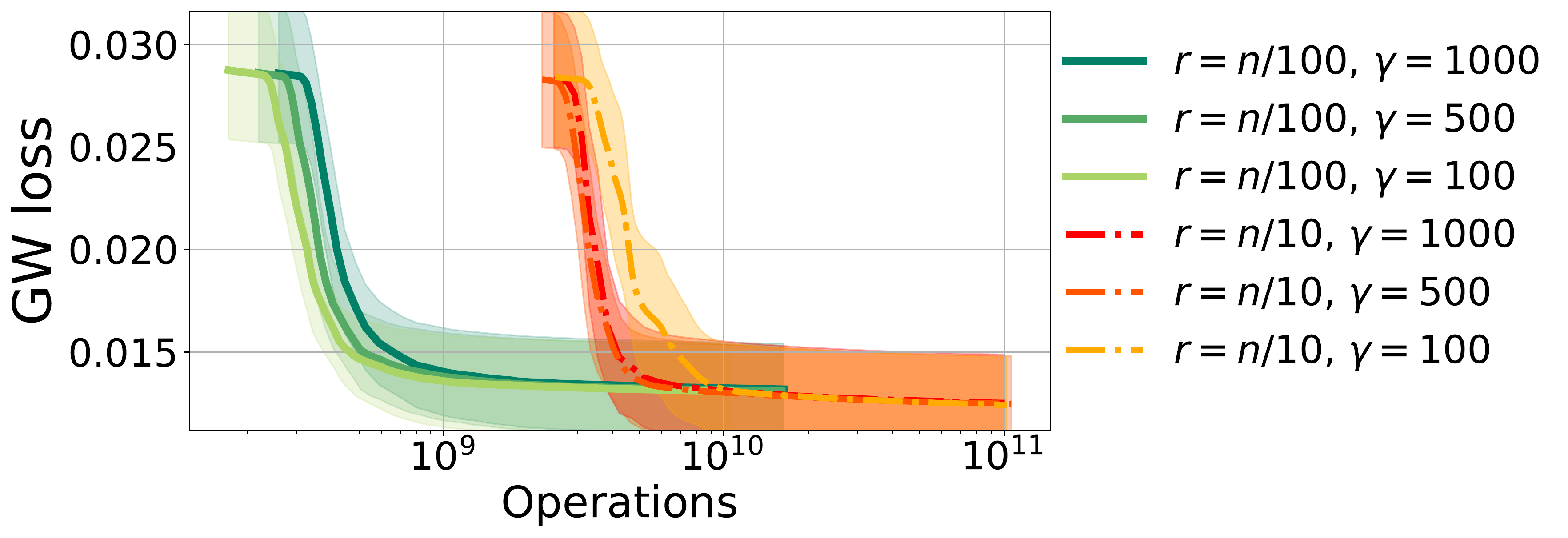}
\caption{We consider two $n=m=1000$ samples of mixtures of (2 and 3) Gaussians in resp. 5 and 10-D, endowed with the squared Euclidean metric, compared with \textbf{Lin LR}. The time/loss tradeoff illustrated in these plots show that our method is only mildly impacted by step size $\gamma$ for both ranks $r=n/100$ and $n/10$. \label{fig-gamma}}
% \vskip-.4cm
\end{figure}
% \end{wrapfigure}

\paragraph{Experiments on Single Cell Genomics Data.} We reproduce the single-cell alignment experiments introduced in~\citep{SCOT2020}. The datasets consist in single-cell multi-omics data generated by co-assays, provided with a ground truth one–to-one correspondence, which can be used to benchmark GW strategies. The SNAREseq dataset~\citep{chen2019high}, with $n=m=1047$ points in $\mathbb{R}^{19}$, describes a real-world experiment; the Splatter dataset~\citep{Zappia133173} with $n=m=5000$ points in $\mathbb{R}^{500}$ is synthetic. We use the pre-processing from~\cite{SCOT2020} to prepare intra-domain distance matrices $\Dista$ and $\Distb$ using a k-NN graph based on correlations, to compute shortest path distances. Note that in that case, one cannot obtain directly in linear time a low-rank factorization of $\Dista$ and $\Distb$ using~\cite{bakshi2018sublinear,indyk2019sampleoptimal}, since the shortest path distances need to be computed first. Therefore, we only use our quadratic approach \textbf{LR} and the cubic implementation of the entropic method \textbf{Ent}, along with \textbf{MREC}. In Fig.~\ref{fig-scot} we compare both the time/GW loss tradeoffs and the alignment performances through the “fraction of samples closer than the true match” (FOSCTTM) error introduced in~\citep{liu2019jointly}. Note that we cannot compare the time-accuracy tradeoff of \textbf{MREC} with our method as the coupling obtained does not satisfy the marginal constraints. \textbf{LR} reaches similar loss, while being orders of magnitude faster than \textbf{Ent}, even for a very small rank $r=n/100$. %Alignments obtained by our method, as scored by FOSCTTM, are robust to the choice of the rank $r$, while other methods are sensitive to hyperparameter choices.

% \begin{figure}
%     \centering
%   \includegraphics[width=0.45\textwidth]{new_figures/plot_acc_vs_time.pdf}
% \caption{In this experiment, we consider the SNAREseq dataset  (\emph{left}) which consists in two point clouds of $n=m=1047$ samples in respectively 10-D and 19-D and the Splatter dataset (\emph{right}) composed of two point clouds of $n=m=5000$ samples in respectively 50-D and 500-D. The cost considered is the shortest path distance and we compare the time-accuracy tradeoffs of our method with the entropic approach. Note that as the coupling obtained by \textbf{MREC} does not satisfy the marginal constraints, we cannot compare directly its accuracy with the other two methods in term of the GW cost. We show that our method reaches similar accuracy while being order of magnitude faster than \textbf{Ent} even for a very small rank $r=n/100$.}\label{fig-scot-acc}
% % \vspace{-0.1cm}
% \end{figure}

% \begin{figure}
%     \centering
%   \includegraphics[width=0.45\textwidth]{new_figures/plot_alignment.pdf}
% \caption{In this experiments we consider the same setting as the one presented in Figure~\ref{fig-scot-acc} and we compare the FOSCTTMs ranked in the increasing order of \textbf{LR}, \textbf{Ent} and \textbf{MREC} when varying their hyperparameters. We see that the alignments obtained by our method are robust to the choice of the rank $r$ and that all the methods manage to reach similar performances.}\label{fig-scot-align}
% % \vspace{-0.1cm}
% \end{figure}

\begin{figure}
% \vspace{-0.3cm}
    \centering
   \includegraphics[width=0.48\textwidth]{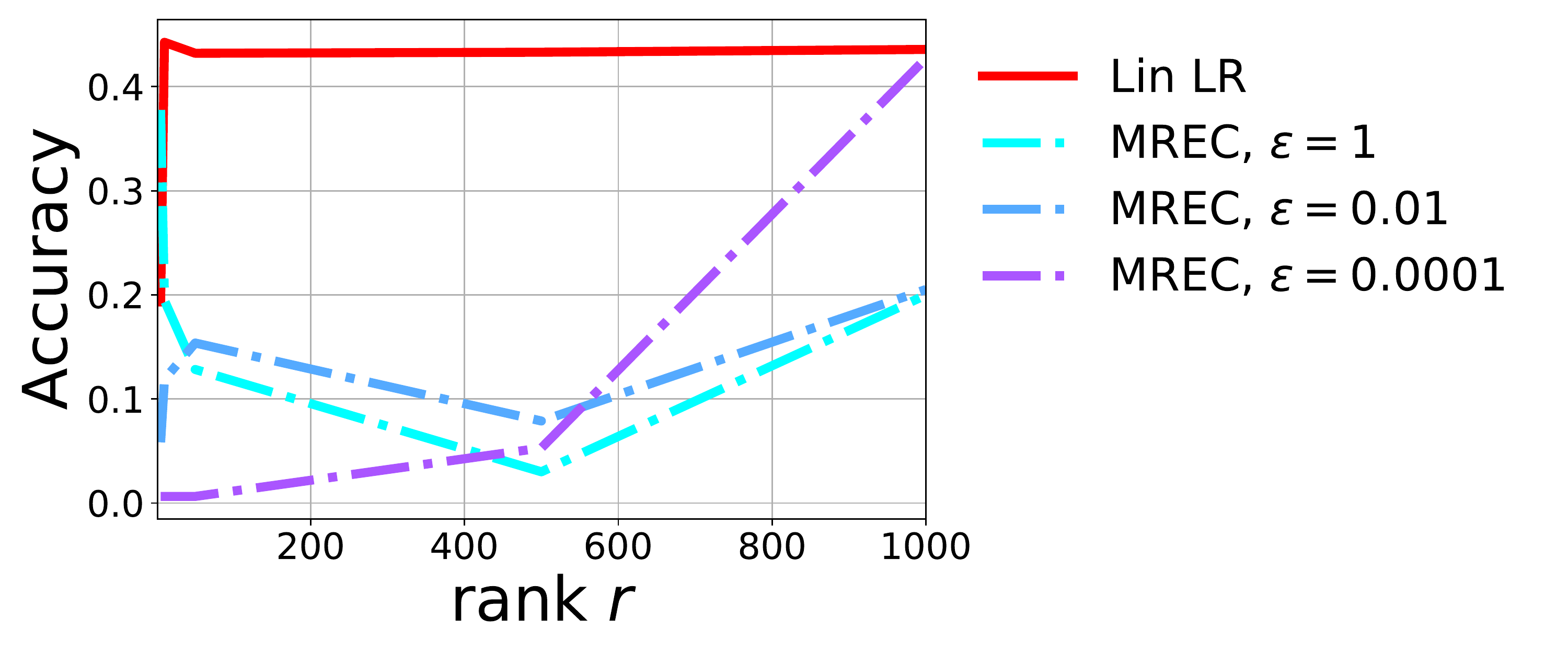}
\caption{Using the BRAIN dataset (two point clouds of $n=34079$ and $m=27906$ samples in 50-D, endowed with squared Euclidean distance) we compare the GW loss  against the rank (or the number of clusters) for both \textbf{Lin LR} and \textbf{MREC} for multiple choices of $\varepsilon$ in \textbf{MREC}. We show that our method is robust to the choice of the rank and obtains consistently better accuracy than \textbf{MREC}.}\label{fig-brain}
% \vspace{-0.1cm}
\end{figure}

\paragraph{Experiment on BRAIN.} We reproduce the experiment proposed in~\cite{blumberg2020mrec}. We consider the dataset introduced in~\cite{lake2018integrative} of single cells sampled from the human brain with eight different cell labels. The dataset contains two groups with different representations: one contains $n=34079$ cells represented by their genes expressions, while the second contains $m=27906$ cells represented by their DNA region accessibilities. We reuse the preprocessing in~\cite{blumberg2020mrec}, by applying the method proposed in~\cite{zheng2017massively} and available in Scanpy~\cite{wolf2018scanpy} to the first group and a TF-IDF representation to the second one. A PCA is then performed on each group to reduce dimensions to 50, endowed with the squared Euclidean distance. These datasets are too large to be handled with entropic approaches, and show the potential of our linear approach \textbf{Lin LR} to handle larger scale problems. To compare \textbf{Lin LR} with \textbf{MREC}, we measure the accuracy of their matchings, as proposed in~\cite{blumberg2020mrec}, by computing the fraction of points in the first group whose associated points under the matching given by the method share the same label in the second group. In Figure~\ref{fig-brain}, we plot the accuracy against the rank (or the number of clusters in MREC) for both \textbf{Lin LR} and \textbf{MREC}. We also consider multiple versions of \textbf{MREC} by varying its entropic regularization parameter, $\varepsilon$, involved in the inner matching of the recursive method. Our method obtains consistently better accuracy than that obtained by \textbf{MREC}. 

% \begin{figure*}[!t]
% \centering
% \includegraphics[width=1\textwidth]{figures/SCOT/plot_accuracy_FOSCTTM_not_all.png}
% \caption{We plot, for each cells of the SNAREseq dataset, the FOSCTTM ranked in the increasing order for both \textbf{GW-LR} and \textbf{Entropic-GW}.}\label{fig-scot-acc-vs-time}
% \vspace{-0.1cm}
% \end{figure*}

% \begin{figure*}[!t]
% \centering
% \includegraphics[width=1\textwidth]{figures/SCOT/plot_accuracy_vs_time_exp2.png}
% \caption{Plot of the time-accuracy tradeoff when varying $\gamma$ for multiple choices of rank $r$ on the SNAREseq dataset. For \textbf{Entropic-GW} we set $\varepsilon=1/\gamma$, for \textbf{GW-LR}, we set $\varepsilon=0$.
% }\label{fig-scot-align}
% \vspace{-0.1cm}
% \end{figure*}

% \begin{figure*}[!t]
% \centering
% \includegraphics[width=1\textwidth]{new_figures/SCOT_splatter.pdf}
% \caption{Here we compare the time-accuracy tradeoffs as well as the FOSCTTMs ranked in the increasing order of the different methods when varying their hyperparameters. \emph{Left, middle-right: } comparison of the time-accuracy tradeoffs between \textbf{GW-LR} and \textbf{Entropic-GW} on respectively the Splatter dataset and the SNAREseq dataset. \emph{Middle-left, right: }
% comparison of the alignment obtained between \textbf{GW-LR}, \textbf{Entropic-GW} and \textbf{MREC}.}\label{fig-scot}
% \vspace{-0.1cm}
% \end{figure*}

\textbf{Discussion.} 
While the factorization introduced in~\cite{scetbon2021lowrank} held the promise to speed up classic OT, we show in this work that it delivers an even larger impact when applied to GW: indeed, the combination of low-rank Sinkhorn factorization with-low rank cost matrices is the only one, to our knowledge, that achieves linear time/memory complexity for the Gromov-Wasserstein problem. The GW problem is NP-hard, its optimal solution out of reach and approximate solutions can only be reached using an inductive bias. Here we propose to compute \textit{efficiently} a coupling whose GW cost is low. By adding low-rank constraints, our goal is no longer to approach the optimal coupling, but rather to promote low-rank solutions among many that have a low GW cost. Our low-rank constraint obtains similar performance as the entropic regularization, the current default approach, while being much faster to compute. We show in experiments that low-rank couplings can reach low GW costs, and that they are directly useful in real-world tasks. Our approach has, however, a few limitations compared to the entropic one: setting $\gamma$, while not problematic in most of our experiments, could require a bit of tuning in order to obtain faster runs in challenging situations. Our assumptions to reach linearity, as discussed in \S\ref{sec:imposing} and \ref{sec-lin-GW} mostly rests on two important assumptions: the rank of distance matrices (the intrisic dimensionality of data points) must be such that $d,d'$ are dominated by $n,m$ and that a small enough rank $r$ be able to capture the configuration of the input measures. Pending these constraints, which are valid in most relevant experimental setups we know of, we have demonstrated that our approach is versatile, remains faithful to the original GW formulation, and scales to sizes that are out of reach for the SoTA entropic solver.

\section*{Acknowledgments}
This work was supported by a "Chaire d’excellence de l’IDEX Paris Saclay", by the French government under management of Agence Nationale de la Recherche as part of the ``Investissements d'avenir'' program, reference ANR19-P3IA-0001 (PRAIRIE 3IA Insti- tute) and by the European Research Council (ERC project NORIA).

%% file: sections/suppmat.tex
\onecolumn
\section*{Supplementary material}
\section{Proofs}
\label{sec-proofs}
\subsection{Proof of Proposition~\ref{prop:init}}
\begin{proof}
Let $(Q,R,g)\in\mathcal{C}(a,b,r,\alpha)$, $P\eqdef Q\Diag(1/g)R^T$. Remarks that for all $i,j$, 
\begin{align*}
    \sqrt{\sum_{i',j'}|\Dista_{i,i'} -\Distb_{j,j'} |^2 P_{i',j'}}&\geq  \left|\sqrt{\sum_{i',j'}|\Dista_{i,i'} |^2 P_{i',j'}} -  \sqrt{\sum_{i',j'}|\Distb_{j,j'} |^2 P_{i',j'}}\right|\\
    &\geq |\sqrt{\tilde{x}_i} - \sqrt{\tilde{y}_j}|
\end{align*}
Therefore we have
\begin{align*}
   \sqrt{\sum_{i,i',j,j'}|\Dista_{i,i'} -\Distb_{j,j'} |^2 P_{i',j'} P_{i,j}}&=\sqrt{\sum_{i,j}\sum_{i',j'}|\Dista_{i,i'} -\Distb_{j,j'} |^2 P_{i',j'} P_{i,j}}\\
   &\geq \sqrt{\sum_{i,j} |\sqrt{\tilde{x}_i} - \sqrt{\tilde{y}_j}|^2P_{i,j}}
\end{align*}
Finally we obtain that
\begin{align*}
  \sum_{i,i',j,j'}|\Dista_{i,i'} -\Distb_{j,j'} |^2 P_{i',j'} P_{i,j} -\varepsilon H(Q,R,g)
   &\geq \sum_{i,j} |\sqrt{\tilde{x}_i} - \sqrt{\tilde{y}_j}|^2P_{i,j} -\varepsilon H(Q,R,g)
\end{align*}
and by taking the infimum over all $(Q,R,g)\in\mathcal{C}(a,b,r,\alpha)$, the results follows.
\end{proof}

\subsection{Proof of Proposition~\ref{prop:cvg-MD-Dykstra}}
To show the result, we first need to recall some notions linked to the relative smoothness. Let $\mathcal{X}$ a closed convex subset in a Euclidean space $\mathbb{R}^q$. Given a convex function $H:\mathcal{X}\rightarrow\mathbb{R}$ continuously differentiable, one can define the \emph{Bregman divergence} associated to $H$ as 
\begin{align*}
    D_H(x,z):=H(x)-H(z)-\langle \nabla H(z),x-z\rangle.
\end{align*}
Let us now introduce the definition of the relative smoothness with respect the $H$.
\begin{defn}[Relative smoothness.]
Let $L>0$ and $f$ continuously differentiable on $\mathcal{X}$. $f$ is said to be $L$-smooth relatively to $H$ if
\begin{align*}
    f(y)\leq f(x) +\langle \nabla f(x),y-x\rangle + L D_H(y,x)
\end{align*}
\end{defn}
In~\citep{scetbon2021lowrank}, the authors show the following general result on the non-asymptotic stationary convergence of the mirror-descent scheme defined by the following recursion:
\begin{align*}
    x_{k+1}=\argmin_{x\in\mathcal{X}} \langle\nabla f(x_k),x\rangle +\frac{1}{\gamma_k}D_h(x,x_k)
\end{align*}
where $(\gamma_k)$ a sequence of positive step-size.
\begin{prop}[\citep{scetbon2021lowrank}]
\label{prop-MD-general}
Let $N\geq 1$, $f$ continuously differentiable on $\mathcal{X}$ which is $L$-smooth relatively to $H$. By considering for all $k=1,\dots, N$, $\gamma_k=1/2L$, and by denoting  $D_0=f(x_0)-\min_{x\in\mathcal{X}}f(x)$, we have
\begin{align*}
    \min_{0\leq k\leq N-1} \Delta_k \leq \frac{4 L D_0}{N}.
\end{align*}
where for all $k=1,\dots, N$
\begin{align*}
   \Delta_k \eqdef \frac{1}{\gamma_k^2}(D_H(x_k,x_{k+1})+D_H(x_{k+1},x_k)).
\end{align*}
\end{prop}

Let us now show that our objective function is relatively smooth with respect the the $\text{KL}$ divergence~\cite{lu2017relativelysmooth,zhang2020wasserstein}. The result of Propostion~\ref{prop:cvg-MD-Dykstra} will then follow from Proposition~\ref{prop-MD-general}. Here $\mathcal{X}=\mathcal{C}(a,b,r,\alpha)$, $H$ is the negative entropy defined as
$$H(Q,R,g):=\sum_{i,j}Q_{i,j}(\log(Q_{i,j})-1)+\sum_{i,j}R_{i,j}(\log(R_{i,j})-1)+\sum_{j}g_{j}(\log(g_{j})-1),$$ and let us define for all $(Q,R,g)\in\mathcal{C}(a,b,r,\alpha) $
\begin{align*}
    F_{\varepsilon}(Q,R,g)\eqdef  -2\langle \Dista Q\Diag(1/g)R^T\Distb, Q\Diag(1/g)R^T\rangle +\varepsilon H(Q,R,g)\;.
\end{align*}
Let us now show the following proposition.
\begin{prop}
Let $\varepsilon\geq 0$, $\frac{1}{r}\geq \alpha> 0$ and let us denote $L_{\varepsilon,\alpha} \eqdef 27(\Vert \Dista\Vert_2\Vert \Distb\Vert_2/\alpha^4+\varepsilon)$. Then for all $(Q_1,R_1,g_1),(Q_2,R_2,g_2)\in \mathcal{C}(a,b,r,\alpha)$, we have
\begin{align*}
     \Vert \nabla F_{\varepsilon}(Q_1,R_1,g_1)-\nabla F_{\varepsilon}(Q_2,R_2,g_2)\Vert_2\leq L_{\varepsilon,\alpha}\Vert H(Q_1,R_1,g_1)- H(Q_2,R_2,g_2)\Vert_2 
\end{align*}
\end{prop}
\begin{proof}
Let $(Q,R,g)\in \mathcal{C}(a,b,r,\alpha)$ and let us denote $P=Q\Diag(1/g)R^T$. We first have that
\begin{align*}
    \nabla F_{\varepsilon}(Q,R,g)=\left(\nabla_Q F_{\varepsilon}(Q,R,g),\nabla_R F_{\varepsilon}(Q,R,g),
    \nabla_g F_{\varepsilon}(Q,R,g)\right)
\end{align*}
where
\begin{align*}
    \nabla_Q F_{\varepsilon}(Q,R,g)&\eqdef -4\Dista P \Distb R\Diag(1/g)+\varepsilon\log Q\\
    \nabla_R F_{\varepsilon}(Q,R,g)&\eqdef -4\Distb P^T \Dista Q\Diag(1/g)+\varepsilon\log R\\
    \nabla_g F_{\varepsilon}(Q,R,g) &\eqdef  -4\mathcal{D}(Q^T\Dista P \Distb R)/g^2+\varepsilon\log g
\end{align*}
First remarks that
\begin{align*}
  \Vert \nabla_Q F_{\varepsilon}(Q_1,R_1,g_1) - \nabla_Q F_{\varepsilon}(Q_2,R_2,g_2) \Vert_2 & \leq 4 \Vert \Dista P_1\Distb R_1 \Diag(1/g_1) - \Dista P_2\Distb R_2 \Diag(1/g_2)\Vert_2\\
  &+ \varepsilon \Vert \log Q_1 - \log Q_2\Vert_2\;.
\end{align*}
Moreover we have
\begin{align*}
\Dista P_1\Distb R_1 \Diag(1/g_1) - \Dista P_2\Distb R_2 \Diag(1/g_2)=
 \Dista ((P_1-P_2) \Distb R_1\Diag(1/g_1) + P_2\Distb( R_1\Diag(1/g_1) - R_2\Diag(1/g_2))
\end{align*} 
where
\begin{align*}
    P_1 - P_2 = (Q_1-Q_2)\Diag(1/g_1) R_1^T + Q_2(\Diag(1/g_1) R_1^T -\Diag(1/g_2) R_2^T)
\end{align*}
and
\begin{align*}
    R_1\Diag(1/g_1) - R_2\Diag(1/g_2) = (R_1 - R_2)\Diag(1/g_1) + R_2 (\Diag(1/g_1)- \Diag(1/g_2))\; .
\end{align*}
Moreover we have 
$$\Vert \Dista P_1\Distb R_1 \Diag(1/g_1) - \Dista P_2\Distb R_2 \Diag(1/g_2)\Vert\leq  \Vert \Dista\Vert \Vert B\Vert \Vert P_1-P_2\Vert/\alpha + \Vert \Dista \Vert \Vert \Distb\Vert \Vert R_1\Diag(1/g_1) - R_2\Diag(1/g_2)\Vert$$
then remark that 
$$ \Vert P_1 - P_2\Vert\leq \Vert Q_1 - Q_2\Vert/\alpha+\Vert R_1\Diag(1/g_1) - R_2\Diag(1/g_2)\Vert$$ and 
$$ \Vert R_1\Diag(1/g_1) - R_2\Diag(1/g_2)\Vert\leq \Vert R_1 - R_2\Vert/\alpha+\Vert 1/g_1-1/g_2\Vert $$ 
from which follows that
\begin{align*}
    \Vert \Dista P_1\Distb R_1 \Diag(1/g_1) - \Dista P_2\Distb R_2 \Diag(1/g_2)\Vert \leq& \frac{\Vert \Dista\Vert \Vert \Distb\Vert}{\alpha}\left(\frac{\Vert Q_1 - Q_2\Vert}{\alpha}+\frac{\Vert R_1 - R_2\Vert}{\alpha}+ \Vert 1/g_1 - 1/g_2\Vert \right) \\
   & + \Vert \Dista\Vert \Vert \Distb\Vert \left(\frac{\Vert R_1 - R_2\Vert}{\alpha}+\Vert 1/g_1 - 1/g_2\Vert\right)\;.
\end{align*}
As $Q\rightarrow H(Q)$ is 1-strongly convex w.r.t to the $\ell_2$-norm on $\Delta_{n\times r}$, we have
\begin{align*}
    \Vert Q_1-Q_2\Vert_2^2&\leq \langle \log Q_1-\log Q_2,Q_1-Q_2\rangle \\
    &\leq \Vert  \log Q_1-\log Q_2\Vert_2\Vert Q_1-Q_2\Vert_2
\end{align*}
from which follows that
\begin{align*}
        \Vert Q_1-Q_2\Vert_2\leq \Vert \log Q_1-\log Q_2\Vert_2.
\end{align*}
Moreover we have
\begin{align*}
    \Vert 1/g_1 - 1/g_2\Vert_2\leq \frac{\Vert g_1-g_2\Vert_2}{\alpha^2}\leq \frac{\Vert \log g_1-\log g_2\Vert_2}{\alpha^2}
\end{align*}
Then we obtain that
\begin{align*}
     \Vert \nabla_Q F_{\varepsilon}(Q_1,R_1,g_1) - \nabla_Q F_{\varepsilon}(Q_2,R_2,g_2) \Vert_2&\leq\left(\frac{4\Vert \Dista\Vert \Vert \Distb\Vert}{\alpha^2} + \varepsilon\right) \Vert \log Q_1-\log Q_2\Vert_2 \\
     &+(1 + 1/\alpha)\frac{4\Vert \Dista\Vert \Vert \Distb\Vert}{\alpha}\Vert \log R_1-\log R_2\Vert_2  \\
      &(1 + 1/\alpha)\frac{4\Vert \Dista\Vert \Vert \Distb\Vert}{\alpha^2} \Vert \log g_1-\log g_2\Vert_2 
\end{align*}
Similarly we obtain that
Then we obtain that
\begin{align*}
     \Vert \nabla_R F_{\varepsilon}(Q_1,R_1,g_1) - \nabla_R F_{\varepsilon}(Q_2,R_2,g_2) \Vert_2&\leq\left(\frac{4\Vert \Dista\Vert \Vert \Distb\Vert}{\alpha^2} + \varepsilon\right) \Vert \log R_1-\log R_2\Vert_2 \\
     &+(1 + 1/\alpha)\frac{4\Vert \Dista\Vert \Vert \Distb\Vert}{\alpha}\Vert \log Q_1-\log Q_2\Vert_2  \\
      &(1 + 1/\alpha)\frac{4\Vert \Dista\Vert \Vert \Distb\Vert}{\alpha^2} \Vert \log g_1-\log g_2\Vert_2 
\end{align*}
Moreover we have
\begin{align*}
   \Vert \nabla_g F_{\varepsilon}(Q_1,R_1,g_1) - \nabla_g F_{\varepsilon}(Q_2,R_2,g_2) \Vert_2 \leq& 4\Vert \mathcal{D}(Q_1^T\Dista P_1 \Distb R_1)/g_1^2 - \mathcal{D}(Q_2^T\Dista P_2 \Distb R_2)/g_2^2\Vert\\
   &+ \varepsilon \Vert \log g_1 - \log g_2\Vert 
\end{align*}
and 
\begin{align*}
\mathcal{D}(Q_1^T\Dista P_1 \Distb R_1)/g_1^2 - \mathcal{D}(Q_2^T\Dista P_2 \Distb R_2)/g_2^2 = &(1/g_1^2-1/g_2^2)\mathcal{D}(Q_1^T\Dista P_1 \Distb R_1) \\
&+\frac{1}{g_2^2} (\mathcal{D}(Q_1^T\Dista P_1 \Distb R_1) - \mathcal{D}(Q_2^T\Dista P_2 \Distb R_2));.
\end{align*}
Note also that 
\begin{align*}
    \Vert (1/g_1^2-1/g_2^2)\mathcal{D}(Q_1^T\Dista P_1 \Distb R_1)\Vert\leq\frac{2\Vert \Dista\Vert \Vert \Distb\Vert}{\alpha^3}\Vert \log g_1 - \log g_2 \Vert 
\end{align*}
and 
\begin{align*}
    Q_1^T\Dista P_1 \Distb R_1 - Q_2^T\Dista P_2 \Distb R_2 &= (Q_1^T-Q_2^T)\Dista P_1 \Distb R_1 + Q_2^T\Dista(P_1 \Distb R_1 - P_2 \Distb R_2)\\
    &=(Q_1^T-Q_2^T)\Dista P_1 \Distb R_1  + Q_2^T\Dista( (P_1-P_2)\Distb R_1 +  P_2\Distb( R_1 - R_2))
\end{align*}
from which follows that
\begin{align*}
\Vert \frac{1}{g_2^2} (\mathcal{D}(Q_1^T\Dista P_1 \Distb R_1) - \mathcal{D}(Q_2^T\Dista P_2 \Distb R_2))\Vert \leq \frac{\Vert \Dista\Vert \Vert \Distb\Vert}{\alpha^2} \left(\Vert \log Q_1 - \log Q_2\Vert + \Vert \log R_1 - \log R_2\Vert + \Vert P_1 - P_2\Vert\right)
\end{align*}
and we obtain that 
\begin{align*}
     \Vert \nabla_g F_{\varepsilon}(Q_1,R_1,g_1) - \nabla_g F_{\varepsilon}(Q_2,R_2,g_2) \Vert_2 &\leq \left(\frac{4\Vert \Dista\Vert \Vert \Distb\Vert}{\alpha^2} + \frac{1}{\alpha}\right) \Vert \log Q_1 - \log Q_2\Vert\\
     &+ \left(\frac{4\Vert \Dista\Vert \Vert \Distb\Vert}{\alpha^2} + \frac{1}{\alpha}\right) \Vert \log R_1 - \log R_2\Vert\\
     &+ \left(\frac{4\Vert \Dista\Vert \Vert \Distb\Vert}{\alpha^4}+\frac{8\Vert \Dista\Vert \Vert \Distb\Vert}{\alpha^3}+\varepsilon\right) \Vert \log g_1 - \log g_2 \Vert 
\end{align*}
Finally we have 
\begin{align*}
     \Vert \nabla F_{\varepsilon}(Q_1,R_1,g_1)-\nabla F_{\varepsilon}(Q_2,R_2,g_2)\Vert_2^2&\leq 3\left[\left(\frac{4\Vert \Dista\Vert \Vert \Distb\Vert}{\alpha^2} + \varepsilon\right)^2 +(1 + 1/\alpha)^2\frac{16\Vert \Dista\Vert^2 \Vert \Distb\Vert^2}{\alpha^2} + \left(\frac{4\Vert \Dista\Vert \Vert \Distb\Vert}{\alpha^2} + \frac{1}{\alpha}\right)^2 \right ] \\
     &\left( \Vert \log Q_1 - \log Q_2\Vert^2 + \Vert \log R_1 - \log R_2\Vert^2\right) \\
     &+ 3\left[2(1 + 1/\alpha)^2\frac{16\Vert \Dista\Vert \Vert^2 \Distb\Vert^2}{\alpha^4} +\left(\frac{4\Vert \Dista\Vert \Vert \Distb\Vert}{\alpha^4}+\frac{8\Vert \Dista\Vert \Vert \Distb\Vert}{\alpha^3}+\varepsilon\right)^2  \right]\\
     &\Vert \log g_1 - \log g_2 \Vert^2
\end{align*}
from which we obtain that 
\begin{align*}
     \Vert \nabla F_{\varepsilon}(Q_1,R_1,g_1)-\nabla F_{\varepsilon}(Q_2,R_2,g_2)\Vert_2^2&\leq L_{\varepsilon,\alpha}^2\left( \Vert \log Q_1 - \log Q_2\Vert^2 + \Vert \log R_1 - \log R_2\Vert^2 + \Vert \log g_1 - \log g_2 \Vert^2\right)
\end{align*}
and the result follows.
\end{proof}

\section{Double Regularization Scheme}
\label{sec:entropic-reg}
Another way to stabilize the method is by considering a double regularization scheme as proposed in~\citep{scetbon2021lowrank} where in addition of constraining the nonnegative rank of the coupling, we regularize the objective by adding an entropic term in $(Q,R,g)$, which is to be understood as that of the values of the three respective entropies evaluated for each term.
\begin{align}
\label{eq-GW-LR-ent}
 \text{GW-LR}^{(r)}_{\varepsilon,\alpha}((a, \Dista),(b, \Distb)) \eqdef\min_{(Q,R,g)\in\mathcal{C}(a,b,r,\alpha)} \mathcal{E}_{\Dista,\Distb}(Q\Diag(1/g)R^T) - \varepsilon H((Q,R,g))\;.
\end{align}

\textbf{Mirror Descent Scheme.} We propose to use a MD scheme with respect to the \text{KL} divergence to approximate $\text{GW-LR}^{(r)}_{\varepsilon,\alpha}$ defined in \eqref{eq-GW-LR-ent}. More precisely, for any $\varepsilon\geq 0$, the MD scheme leads for all $k\geq 0$ to the following updates which require solving a convex barycenter problem per step:
\begin{equation}
\label{eq-barycenter-GW-inner-ent}  
 (Q_{k+1},R_{k+1},g_{k+1}) \eqdef \!\! \argmin_{\zeta \in\mathcal{C}(a,b,r,\alpha)} \!\! \text{KL}(\zeta,K_k)
\end{equation}
where $(Q_0,R_0,g_0)\in\mathcal{C}(a,b,r)$ is an initial point such that $Q_0>0$ and $R_0>0$, $P_k\eqdef Q_k\Diag(1/g_k)R_k^T$, 
$K_k \eqdef (K_{k}^{(1)},K_{k}^{(2)},K_{k}^{(3)})$, $K_{k}^{(1)} \eqdef \exp(4\gamma \Dista P_k\Distb R_k\Diag(1/g_k) -(\gamma\varepsilon-1)\log(Q_k))$, $K_{k}^{(2)} \eqdef \exp(4\gamma \Distb P^T_kD Q_k \Diag(1/g_k)- (\gamma\varepsilon-1)\log(R_k))$,
$K_{k}^{(3)} \eqdef \exp(-4\gamma\omega_k/g_k^2- (\gamma\varepsilon-1)\log(g_k))$ with  $[\omega_k]_i \eqdef [Q_k^T\Dista P_k\Distb R_k]_{i,i}$ for all $i\in\{1,\dots,r\}$ and $\gamma$ is a positive step size. Solving~\eqref{eq-barycenter-GW-inner} can be done efficiently thanks to the Dykstra’s Algorithm as showed in~\citep{scetbon2021lowrank}. See Appendix~\ref{sec-Dykstra-app} for more details.

\textbf{Convergence of the mirror descent.} Even if the objective~(\ref{eq-GW-LR-ent}) is not convex in $(Q,R,g)$, we obtain the non-asymptotic stationary convergence of the MD algorithm in this setting. 
For that purpose we consider the same convergence criterion as the one proposed in~\citep{scetbon2021lowrank} to obtain non-asymptotic stationary convergence of the MD scheme defined as
\begin{align*}
   \Delta_{\varepsilon,\alpha}(\xi,\gamma) \eqdef \frac{1}{\gamma^2}(\text{KL}(\xi,\mathcal{G}_{\varepsilon,\alpha}(\xi,\gamma))+\text{KL}(\mathcal{G}_{\varepsilon,\alpha}(\xi,\gamma),\xi))
\end{align*}
where $\mathcal{G}_{\varepsilon,\alpha}(\xi,\gamma) \eqdef \argmin_{\zeta\in \mathcal{C}(a,b,r,\alpha)}\{ \langle \nabla \mathcal{E}_{\Dista,\Distb}(\xi),\zeta \rangle +\frac{1}{\gamma} \text{KL}(\zeta,\xi) \}.$
For any $1/r\geq \alpha> 0$, we show in the following proposition the non-asymptotic stationary convergence of the MD scheme applied to the problem~(\ref{eq-GW-LR-ent}). See Appendix~\ref{sec-proofs} for the proof.
\begin{prop}
\label{prop:cvg-MD-Dykstra-ent}
Let $\varepsilon\geq 0$, $\frac{1}{r}\geq \alpha> 0$ and $N\geq 1$. By denoting $L_{\varepsilon,\alpha} \eqdef 27(\Vert \Dista\Vert_2\Vert \Distb\Vert_2/\alpha^4+\varepsilon)$ and by considering a constant stepsize in the MD scheme~(\ref{eq-barycenter-GW-inner}) $\gamma=\frac{1}{2L_{\varepsilon,\alpha}}$, we obtain that
\begin{align*}
    \min_{1\leq k\leq N}\Delta_{\varepsilon,\alpha}((Q_k,R_k,g_k),\gamma)\leq \frac{4L_{\varepsilon,\alpha}  D_0}{N}.
\end{align*}
where $D_0 \eqdef  \mathcal{E}_{\Dista,\Distb}(Q_0\Diag(1/g_0R_0^T)- \text{\normalfont{GW-LR}}^{(r)}((a, \Dista),(b, \Distb))$ is the distance of the initial value to the optimal one. 
\end{prop}

\section{Low-rank Approximation of Distance Matrices}
\label{sec-lr-cost-mat}
Here we recall the algorithm used to perform a low-rank approximation of a distance matrix~\cite{bakshi2018sublinear,indyk2019sampleoptimal}. We use the implementation of~\cite{scetbon2021lowrank}. 
\begin{algorithm}[H]
\SetAlgoLined
\textbf{Inputs:} $X,Y,r,\gamma$\\
Choose $i^{*}\in\{1,\dots,n\}$, and $j^*\{1,\dots,m\}$ uniformly at random.\\
For $i=1,\dots,n$,~$p_i\gets d(x_i,y_j^*)^2 +  d(x_i^*,y_j^*)^2+\frac{1}{m}\sum_{j=1}^m d(x_i^*,y_j)^2$.\\
Independently choose $i^{(1)},\dots,i^{(t)}$ according $(p_1,\dots,p_n)$.\\
$X^{(t)}\gets [x_{i^{(1)}},\dots,x_{i^{(t)}}],~P^{(t)}\gets [\sqrt{t p_{i^{(1)}}},\dots,\sqrt{t p_{i^{(t)}}}],~S\gets d(X^{(t)},Y)/P^{(t)}$\\
Denote $S=[S^{(1)},\dots,S^{(m)}]$,\\
For $j=1,\dots,m$,~$q_j\gets \Vert S^{(j)} \Vert_2^2/\Vert S\Vert_F^2$\\
Independently choose $j^{(1)},\dots,j^{(t)}$ according $(q_1,\dots,q_m)$.\\
$S^{(t)}\gets  [S^{j^{(1)}},\dots,S^{j^{(t)}}],~Q^{(t)}\gets [\sqrt{t q_{j^{(1)}}},\dots,\sqrt{t q_{j^{(t)}}}],~W\gets S^{(t)}/Q^{(t)}$\\
$U_1,D_1,V_1 \gets \text{SVD}(W)$ (decreasing order of singular values).\\
$N\gets [U_1{(1)},\dots,U_1^{(r)}],~N\gets S^T N/\Vert W^T N\Vert_F$\\
Choose $j^{(1)},\dots,j^{(t)}$ uniformly at random in $\{1,\dots,m\}$.\\
$Y^{(t)}\gets [y_{j^{(1)}},\dots,y_{j^{(t)}}], D^{(t)}\gets d(X,Y^{(t)})/\sqrt{t}$.\\
$U_2,D_2,V_2 = \text{SVD}(N^T N),~U_2\gets U_2/D_2,~N^{(t)} \gets [(N^T)^{(j^{(1)})},\dots,(N^T)^{(j^{(t)})}] ,~B\gets U_2^T N^{(t)}/\sqrt{t},~A\gets (BB^T)^{-1}$.\\
$Z\gets AB(D^{(t)})^T,~M\gets Z^{T}U_2^T$\\
\textbf{Result:} $M,N$
\caption{$\text{LR-Distance}(X,Y,r,\gamma)$~~\cite{bakshi2018sublinear,indyk2019sampleoptimal} \label{alg-LR-distance}}
\end{algorithm}

\section{Nonnegative Low-rank Factorization of the Couplings}
\label{sec-Dykstra-app}
In this section, we recall the algorithm presented in~\citep{scetbon2021lowrank} to solve problem~\eqref{eq-barycenter-GW-inner} where we denote $p_1\eqdef a$ and  $p_2\eqdef b$.

\begin{algorithm}
\SetAlgoLined
\textbf{Inputs:} $K^{(1)},K^{(2)},\tilde{g} \eqdef K^{(3)},p_1,p_2,\alpha,\delta,q^{(3)}_1=q^{(3)}_2=\mathbf{1}_r,\forall i\in\{1,2\},~ \tilde{v}^{(i)}=\mathbf{1}_r, q^{(i)}=\mathbf{1}_r$\\
\Repeat{$\sum_{i=1}^2\|u^{(i)}\odot K^{(i)}v^{(i)} - p_i\|_1 <\delta$}
{
    $u^{(i)}\gets p_i/K^{(i)}\tilde{v}^{(i)}~\forall i\in\{1,2\}$,
    
    $g\gets \max(\alpha,\tilde{g}\odot q^{(3)}_1),~q^{(3)}_1\gets (\tilde{g}\odot q^{(3)}_1)/ g,~\tilde{g}\gets g$,
    
    $g\gets (\tilde{g}\odot q^{(3)}_2)^{1/3} \prod_{i=1}^2 (v^{(i)}\odot q^{(i)}\odot(K^{(i)})^Tu^{(i)})^{1/3}$,
    
    $v^{(i)}\gets g/(K^{(i)})^T u^{(i)} ~\forall i\in\{1,2\}$,
    
    $q^{(i)}\gets (\tilde{v}^{(i)}\odot q^{(i)})/v^{(i)}~\forall i\in\{1,2\},~q^{(3)}_2 \gets (\tilde{g}\odot q^{(3)}_2)/g$,
    
    $\tilde{v}^{(i)}\gets v^{(i)}~\forall i\in\{1,2\},~\tilde{g}\gets g$,
}
    
$Q\gets  \Diag(u^{(1)})K^{(1)} \Diag(v^{(1)})$\\
$R\gets  \Diag(u^{(2)})K^{(2)} \Diag(v^{(2)})$\\
\textbf{Result:} $Q,R,g$
\caption{$\text{LR-Dykstra}((K^{(i)})_{1\leq i\leq 3},p_1,p_2,\alpha,\delta)$~~\citep{scetbon2021lowrank}}
\end{algorithm}

\section{Additional Experiements}

\subsection{Illustration}
\label{sec-illusation-fig1}
\begin{figure*}[!h]
\centering
\includegraphics[width=1\textwidth]{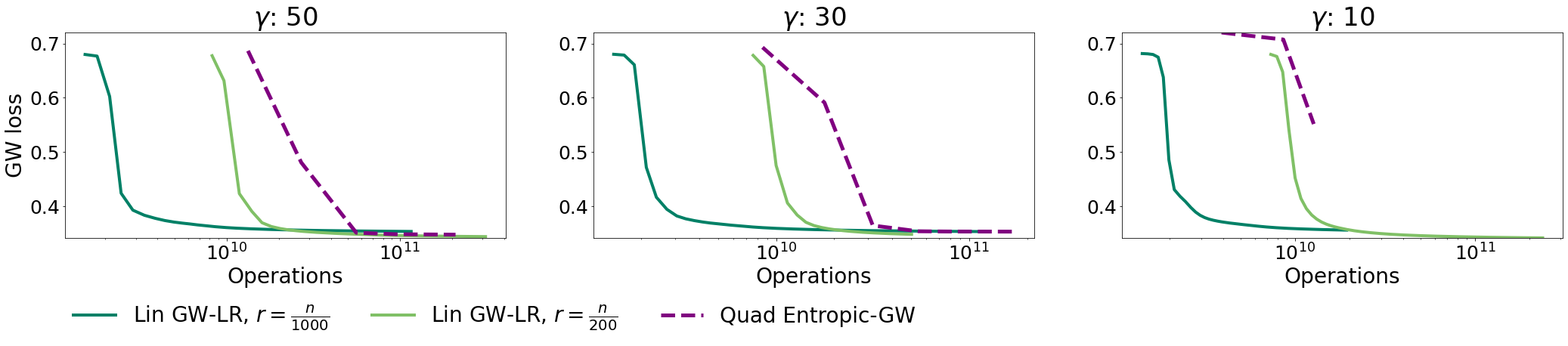}
\caption{Here $n=m=10000$, and the ground cost considered is the squared Euclidean distance. Note that for in that case we have an exact low-rank factorization of the cost. Therefore we compare only \textbf{Quad Entropic-GW} and \textbf{Lin GW-LR}. We plot the time-accuracy tradeoff when varying $\gamma$ for multiple ranks $r$. $\varepsilon=1/\gamma$ for \textbf{Quad Entropic-GW} and $\varepsilon=0$ for \textbf{Lin GW-LR}.
}\label{fig-GT-acc-vs-time}
\vspace{-0.1cm}
\end{figure*}

In Fig.~\ref{fig-GT-acc-vs-time}, we show the time-accuracy tradeoffs of the two methods presented in Figure~\ref{fig-GT-coupling-1} on the same example. We see that our method, \textbf{Lin GW-LR}, manages to obtain similar accuracy as the one obtained by \textbf{Quad Entropic-GW} even when the rank $r=n/1000$ while being much faster with order of magnitude.

\subsection{Effect of $\gamma$ and $\alpha$}
 \label{sec-hyperparam}
\begin{figure}[h!]
\centering
\includegraphics[width=0.4\textwidth]{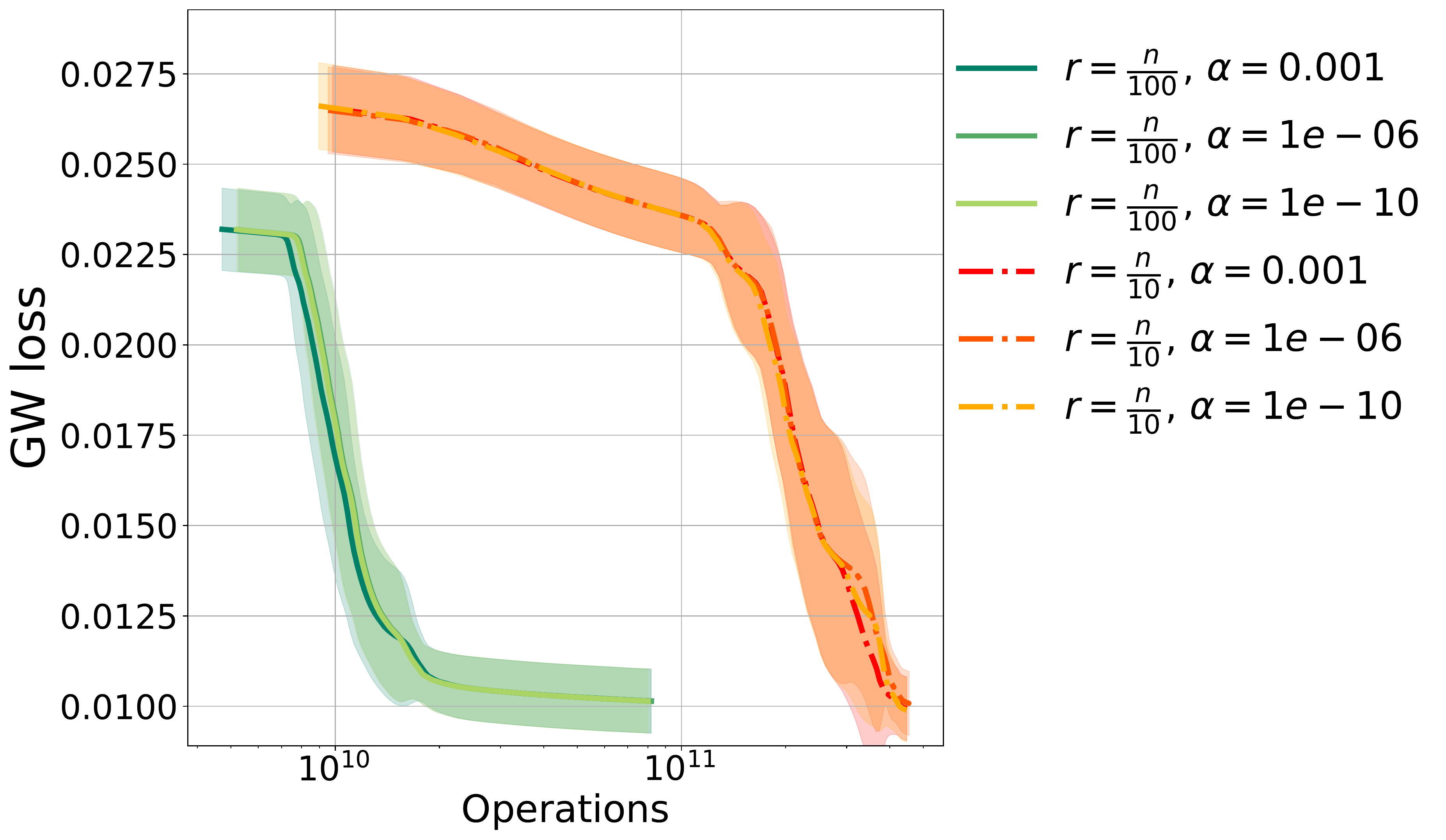}
\caption{We consider $n=m=5000$ samples of mixtures of (2 and 3) Gaussians in resp. 5 and 10-D, endowed with the squared Euclidean metric, compared with \textbf{Lin LR}. The time/loss tradeoff illustrated in these plots show that our method is not impacted by step size $\alpha$ for both ranks $r=n/100$ and $n/10$. }\label{fig-alpha}
\end{figure}

In Fig.~\ref{fig-gamma} and~\ref{fig-alpha}, we consider two Gaussian mixture densities in respectively 5-D and 10-D where we generate randomly the mean and covariance matrice of each Gaussian density using the wishart distribution.

% \begin{align*}
%     &\mu_{\mathcal{X}}^{(1)} = [0,0]\in\mathbb{R}^2,~\mu_{\mathcal{X}}^{(2)}=[0,1]\in\mathbb{R}^2,~\mu_{\mathcal{X}}^{(3)}=[1,1]\in\mathbb{R}^2,\\
%     &~\nu_{\mathcal{Y}}^{(1)} = [0.5,0.5,0]\in\mathbb{R}^{3},~\nu_{\mathcal{Y}}^{(2)}=[-0.5,0.5,0]\in\mathbb{R}^{3},\\
%     &\Sigma_{\mathcal{X}} =0.05\times\text{Id}_{2} \text{\quad and\quad } \Sigma_{\mathcal{Y}} =0.05\times\text{Id}_{3}.
% \end{align*} 
 
\subsection{Effect of the Rank}
\label{sec-effect-rank}
\begin{figure}[h!]
\centering
\includegraphics[width=0.4\textwidth]{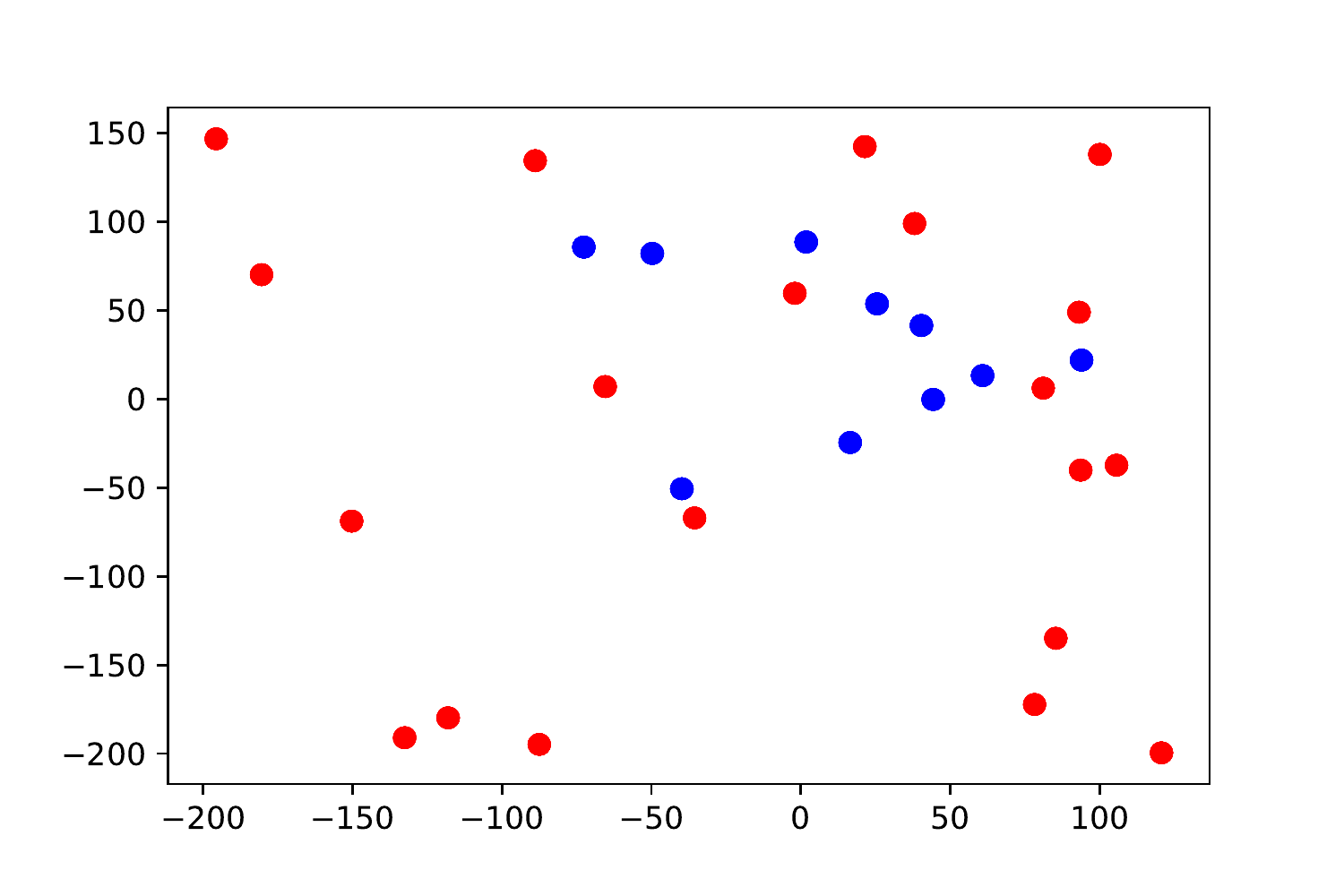}
\caption{We consider two isotropic Gaussian blobs with respectively 10 and 20 centers in 10-D and 15-D and $n=m=5000$ samples and we plot their 2 first coordinates.}\label{fig-blobs}
\end{figure}

In this experiment we compare two isotropic Gaussian blobs with respectively 10 and 20 centers in 10-D and 15-D and $n=m=5000$ samples. In Fig.~\ref{fig-blobs}, we show the two first coordinates of the dataset considered.

\subsection{Low-rank Problem} 
\label{sec-low-rank-problem}

\begin{figure*}[!h]
\centering
\includegraphics[width=0.5\textwidth]{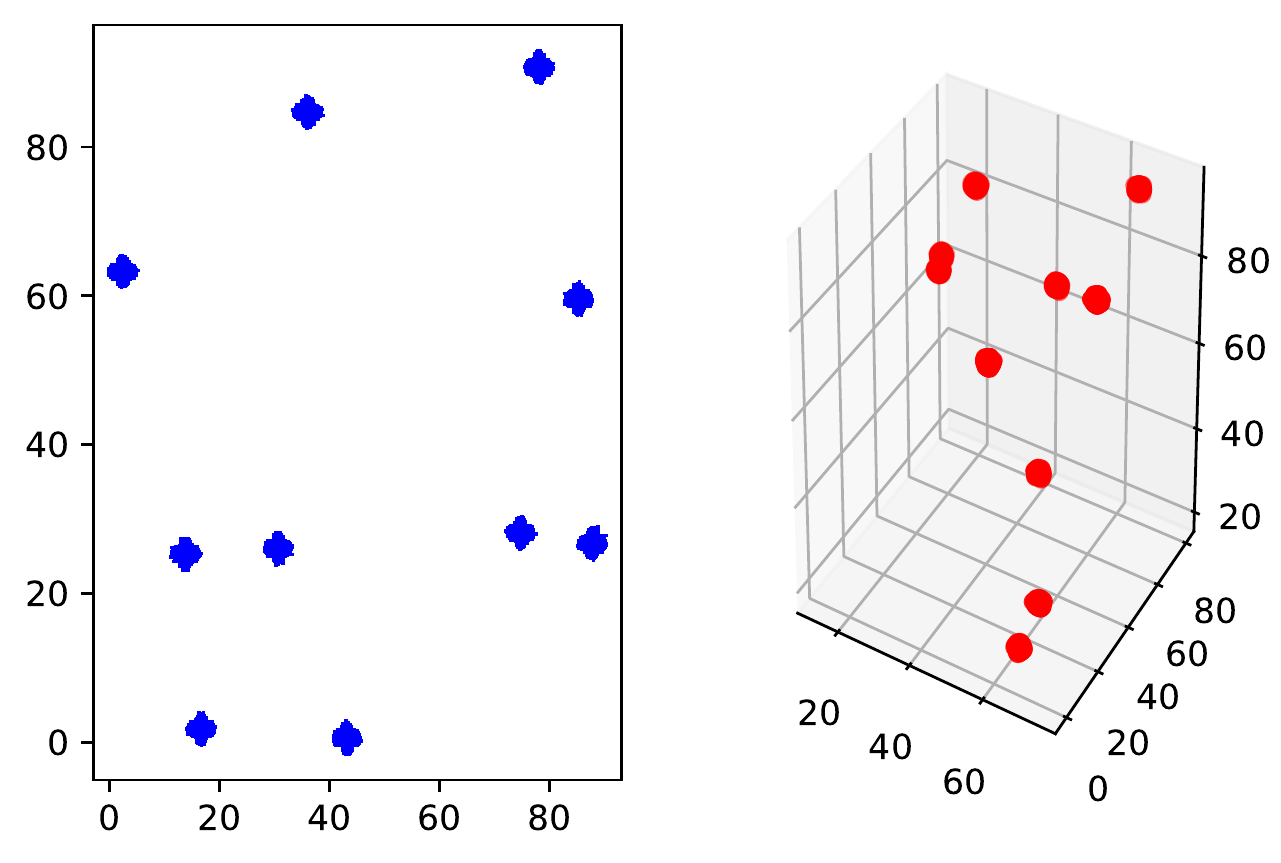}
\caption{The source distribution and the target distribution live respectively in $\mathbb{R}^2$ and $\mathbb{R}^3$. Both distributions have the same number of samples $n=m=10000$, the same number of clusters which is set to be $10$ here, the same number of points in each cluster, and we force the distance between the centroids of the cluster to be larger than $\beta=10$ in each distribution.
}\label{fig-data-bloc}
\vspace{-0.1cm}
\end{figure*}

In Fig.~\ref{fig-bloc-Euclidean}, \ref{fig-bloc-Euclidean-factorized} and~\ref{fig-bloc-SEuclidean}, we consider two distributions in respectively 10-D and 15-D where the support is a concatenation of clusters of points. In Fig.~\ref{fig-data-bloc}, we show an illustration of the distributions considered in smaller dimensions.

\subsection{Ground Truth Experiment}
\begin{figure*}[!h]
\centering
\includegraphics[width=1\textwidth]{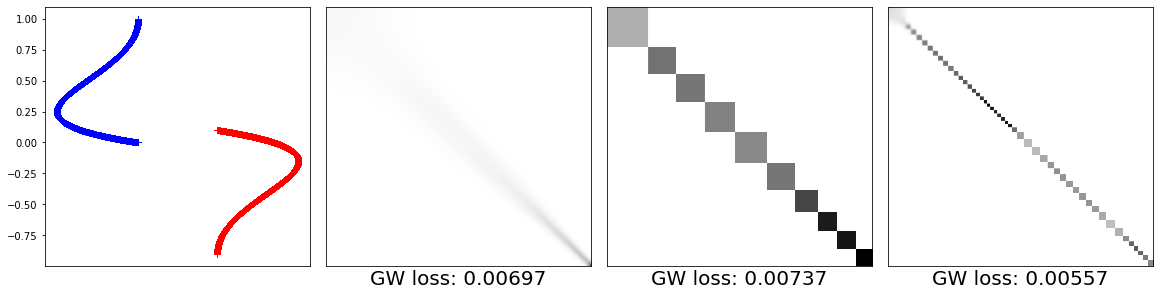}
\caption{We compare the couplings obtained when the ground truth is the identity matrix in the same setting as in Figure~\ref{fig-GT-acc-vs-time}. Here the comparison is done when $\gamma=250$. \emph{Left:} illustration of the dataset considered. \emph{Middle left:} we show the coupling as well as the GW loss obtained by \textbf{Quad Entropic-GW}. \emph{Middle right, right:} we show the couplings and the GW losses obtained by \textbf{Lin GW-LR} when the rank is respectively $r=10$ and $r=100$.
}\label{fig-GT-coupling-id}
\vspace{-0.1cm}
\end{figure*}

\begin{figure*}[!h]
\centering
\includegraphics[width=1\textwidth]{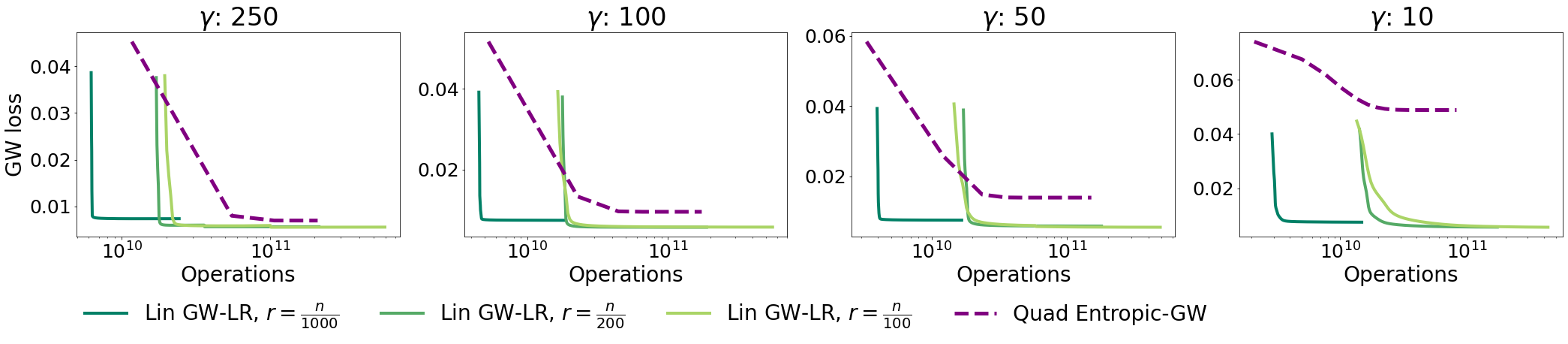}
\caption{The ground truth here is the identity matrix and the true GW loss to achieve is 0. We set the number of samples to be $n=m=10000$. As we consider the squared Euclidean distance, only \textbf{Quad Entropic-GW} and \textbf{Lin GW-LR} are compared. We plot the time-accuracy tradeoff when varying $\gamma$ for multiple choices of rank $r$. $\varepsilon=1/\gamma$ for \textbf{Quad Entropic-GW} and $\varepsilon=0$ for \textbf{Lin GW-LR}.
}\label{fig-GT-acc-vs-time-id}
\vspace{-0.1cm}
\end{figure*}

In this experiment we aim at comparing the different methods when the optimal coupling solving the GW problem has a full rank. For that purpose we consider a certain shape in 2-D which corresponds to the support of the source distribution and we apply two isometric transformations to it, which are a rotation and a translation to obtain the support the target distribution. See Figure~\ref{fig-GT-coupling-id}~(\emph{left}) for an illustration of the dataset. Here we set $a$ and $b$ to be uniform distributions and the underlying cost is the squared Euclidean distance. Therefore the optimal coupling solution of the GW problem is the identity matrix and the GW loss must be 0. In Figure~\ref{fig-GT-acc-vs-time-id}, we compare the time-accuracy tradeoffs, and we show that even in that case, our methods obtain a better time-accuracy tradeoffs for all $\gamma$. See also Figure~\ref{fig-GT-coupling-id} for a comparison of the couplings obtained by the different methods.

